\documentclass[10pt,twocolumn,letterpaper]{article}

\usepackage{cvpr}              % To produce the CAMERA-READY version
\usepackage{amsfonts}
\usepackage{wrapfig}

\definecolor{cvprblue}{rgb}{0.21,0.49,0.74}
\usepackage[pagebackref,breaklinks,colorlinks,allcolors=cvprblue]{hyperref}
\usepackage{wrapfig}

\def\paperID{2551} % *** Enter the Paper ID here
\def\confName{CVPR}
\def\confYear{2026}

\title{3D Gaussian Splatting with Self-Constrained Priors \\ 
for High Fidelity Surface Reconstruction}

\author{%
{Takeshi Noda}$^1$
\quad {Yu-Shen Liu}$^1$$^\dagger$ \ \ \quad {Zhizhong Han}$^2$\     \\
  $^1$School of Software, Tsinghua University, Beijing, China \\
  $^2$Department of Computer Science, Wayne State University, Detroit, USA \\
  \tt\small{yeth24@mails.tsinghua.edu.cn}\quad 
  \tt\small{liuyushen@tsinghua.edu.cn} \quad 
  \tt\small{h312h@wayne.edu}
  }
\begin{document}
\maketitle

\begin{abstract}
Rendering 3D surfaces has been revolutionized within the modeling of radiance fields through either 3DGS or NeRF. Although 3DGS has shown advantages over NeRF in terms of rendering quality or speed, there is still room for improvement in recovering high fidelity surfaces through 3DGS. To resolve this issue, we propose a self-constrained prior to constrain the learning of 3D Gaussians, aiming for more accurate depth rendering. Our self-constrained prior is derived from a TSDF grid that is obtained by fusing the depth maps rendered with current 3D Gaussians. The prior measures a distance field around the estimated surface, offering a band centered at the surface for imposing more specific constraints on 3D Gaussians, such as removing Gaussians outside the band, moving Gaussians closer to the surface, and encouraging larger or smaller opacity in a geometry-aware manner. More importantly, our prior can be regularly updated by the most recent depth images which are usually more accurate and complete. In addition, the prior can also progressively narrow the band to tighten the imposed constraints. We justify our idea and report our superiority over the state-of-the-art methods in evaluations on widely used benchmarks. Project page: \url{ https://takeshie.github.io/GSPrior/}
\renewcommand{\thefootnote}{}
\footnote{$^\dagger$ Corresponding author.}
\end{abstract}

\section{Introduction}
3D Gaussian splatting (3DGS)~\cite{kerbl20233d} becomes a crucial tool for novel view synthesis. Learning explicit 3D Gaussian functions with attributes like color and opacity, 3DGS represents scenes as radiance fields that can be rendered into RGB images with a differentiable splatting operation. Compared with ray tracing based neural rendering methods~\cite{mildenhall2020nerf,deng2022dsnerf,wang2022hf-neus}, 3DGS shows faster rendering speed and better visual fidelity. However, it is still limited in recovering accurate geometry.

Recent works on improving accuracy of depth rendering adopts different strategies, such as reformulating rendering equations for unbiased depth inference~\cite{chen2024vcr,xu2025depthsplat}, imposing multi-view consistency constraints on rendered depth~\cite{chen2024pgsr,zhang2025materialrefgs,chen2024mvsplat}, constraining Gaussian movements within a distance field~\cite{li2025gaussianudf,zhang2024gspull}, or getting supervised by additional priors from data-driven based models~\cite{li2025va,jiang2025gausstr,zuo2025fmgs}.
However, these strategies struggle to add constraints either directly on 3D Gaussians or adaptively in a geometry-aware manner. Without explicit 3D supervision, previous methods are limited in recovering geometry details, and rely on geometric assumptions or pretrained priors which usually do not generalize well on complex or unseen scenes, leading to artifacts and degraded geometric fidelity.

To resolve this issue, we propose a self-constrained prior to impose constraints on 3D Gaussians in a geometry-aware manner. Our prior is derived from depth maps rendered by the 3D Gaussians that are optimized in the radiance field without external knowledge. Along with the prior, we further introduce a coarse-to-fine strategy to progressively refine the prior with the most current depth rendering that turns out to be more accurate. Specifically, our prior is derived from the TSDF grid fused by depth maps, which measures the distance field around the estimated surface. In the grid, we define a band centered at the surfaces, where we remove Gaussians outside the band and impose constraints on the opacity of Gaussians within the band to better align with the current surface estimation. The geometry-aware constraints reduce the impact brought by Gaussian outliers, and make the constraints become targeted on specific Gaussians. Moreover, the refining strategy not only refines the prior but also provides a way to progressively adjust the band to tighten the imposed constraints. We conduct evaluations on widely used benchmark to justify our idea and report our superiority over the state-of-the-art methods. Our contributions are listed below,

\begin{itemize}
\item We propose a self-constrained prior to impose constraints on the learning of 3D Gaussians in a geometry-aware manner.
\item We introduce a novel framework and losses to work with the self-constrained prior to recover more accurate geometry through 3DGS.
\item We report state-of-the-art results in accuracy on the widely used benchmarks.
\end{itemize}
% \section{Introduction}
%-------------------------------------------------------------------------
\begin{figure*}
    \centering
    \includegraphics[width=\textwidth]{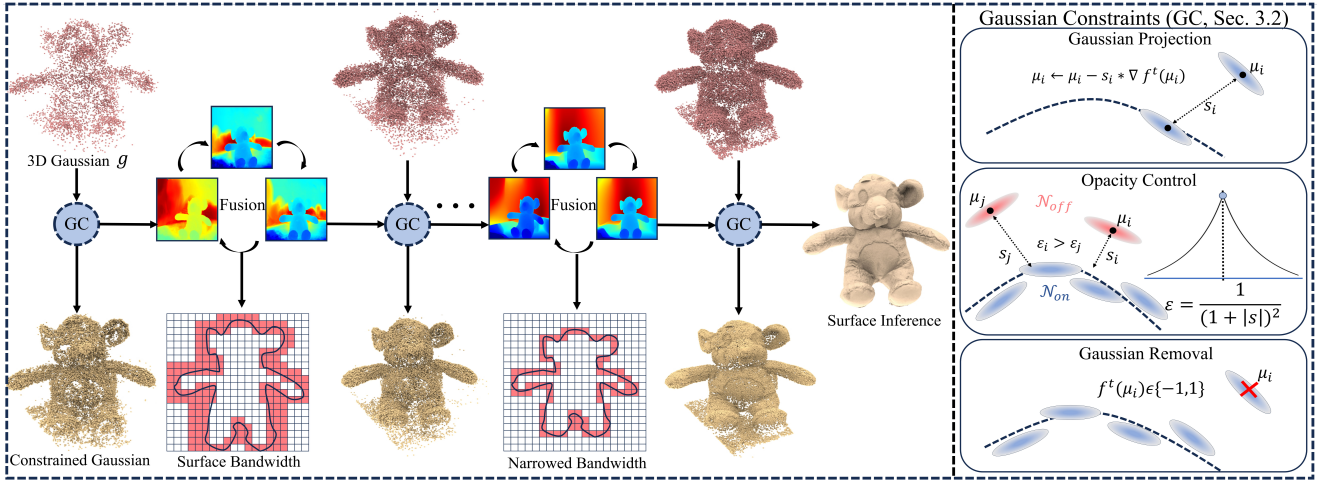}

    \caption{Overview of our method. Given 3D Gaussians $g$, we employ a distance field specified by a fused TSDF grid as our prior $f^{t}$. With $f^{t}$, we define a bandwidth by the surface and iteratively refine $f^{t}$ with updated depth renderings. We also apply Gaussian geometric constraints $(\mathrm{GC})$ that are related to interpolated distance $s$, centers $\mu$ and gradients $\nabla f^{t}$ for high fidelity surface reconstruction.
}
    \label{fig:main_pic}
\end{figure*}
\vspace{0.5cm}
\section{Related Work}
\textbf{Surface Reconstruction with Multi-view Images. }Neural Radiance Fields (NeRFs) \cite{mildenhall2020nerf} have shown promising breakthroughs in novel view synthesis \cite{garbin2021fastnerf,gu2023ue4,chen2023mobilenerf,huang2024neusurf,cao2023hexplane,liao2025litegs} and surface reconstruction \cite{azinovic2022neuralrgbd,huang2024neusurf,deng2022dsnerf,wang2022hf-neus,fu2022geoneus,long2023neuraludf,zhou2026udfstudio}. Previous methods model continuous density and color fields with the differentiable volume-rendering and implicit representations \cite{long2023neuraludf, meng2023neat,oechsle2021unisurf,tang2024nd}. With learned implicit fields, the Marching Cubes \cite{Lorensen87marchingcubes} algorithm is utilized to extract the zero-level-set as the surface. As data scale grows, the NeRFs based methods face significant challenges in training and inference efficiency. To address this, some studies introduce sparse or explicit grid structures \cite{sun2022direct,chen2022tensorf,fridovich2022plenoxels} , multi-resolution representations \cite{muller2022instant,li2023neuralangelo}, and geometric priors \cite{zhang2022critical, fu2022geoneus, wang2022neuris,liang2023helixsurf,zhang2026vrp}. However, these strategies still struggle to balance training efficiency and reconstruction fidelity. 3DGS based methods \cite{kerbl20233d,liu2026speeding,ren2025fastgs,liu2026speeding} achieves superior efficiency in novel view synthesis and geometry modeling with explicit 3D Gaussians. Recent studies \cite{yu2024gsdf,dai2024high,huang20242dgs,wolf2024gs2mesh,wu2024surface,fang2026more,gao2025anisdf,dens3r,MoRE2025} further extend the framework to recover surfaces or geometries. Typically, 2DGS \cite{huang20242dgs} and GSurfel \cite{dai2024high} constrains Gaussians to 2D surfels for better surface approximation. However, these methods still struggle to recover geometry details.\\
\textbf{Learning Gaussians with Implicit Representations. }Current Gaussian-based methods \cite{huang20242dgs,chen2024pgsr,huang2025fatesgs,kerbl20233d,dai2024high} construct TSDFs from depth maps but produce discontinuous surfaces at regions with complex geometry. Meanwhile, surface inference without explicit 3D constraints fails to recover accurate geometry. To solve this issue, recent studies employ implicit field representations \cite{zhang2024gspull,yu2024gsdf,lyu20243dgsr}. GSPull \cite{zhang2024gspull} progressively pulls Gaussians toward object surfaces to build a continuous implicit representation. Similarly, GS-UDF \cite{li2025gaussianudf} additionally learns an unsigned distance field (UDF) to project Gaussians with implicit surfaces for accurate reconstruction in open-surface scenarios. GOF \cite{yu2024gaussian} introduces a volumetric renderer explicitly models opacity and employs Marching Tetrahedra \cite{shen2021deep} for adaptive meshing. To combine the strengths of Gaussians and implicit fields, GSDF \cite{yu2024gsdf} jointly learns both representations and enforces mutual supervision with depth and normal cues for better reconstruction. However, the joint optimization strategy requires a differentiable formulation between the radiance field and the implicit function, which complexes the framework.\\
% \begin{wrapfigure}{l}{0.5\linewidth}
% \centering
% % \vspace{-0.1in}
% \includegraphics[width=\linewidth]{figs/Curves.pdf}\hspace*{-0.3cm}
% \vspace{-0.1in}
%     \caption{Illustration of bandwidth and the range of opacity control.}
%     \label{fig:curves}
%     \vspace{-0.18in}
% \end{wrapfigure} 
\noindent\textbf{Inferring Surfaces through 3DGS. }Enhancing geometry through priors has been widely explored in recent studies \cite{wu2024surface,guedon2024sugar,zhang2024rade,chen2024vcr,li2025va,li2025geosvr,turkulainen2024dnsplatter,li2025va,wolf2024gs2mesh}. PGSR \cite{chen2024pgsr} incorporates multi-view constraints to improve both rendering quality and geometric fidelity. However, local over-smoothing is observed in the reconstructed surfaces, as the 3D Gaussians are not sufficiently constrained by image-based supervision. To further enhance reconstruction, some methods introduce geometric priors for supervision. DN-Splatter \cite{turkulainen2024dnsplatter} leverages depth and normal priors as geometric constraints during optimization. Similarly, SuGaR \cite{guedon2024sugar} builds density fields with a regularization term to align Gaussians with the scene and extracts surfaces via Poisson reconstruction \cite{kazhdan2013screened}. However, geometric and implicit field priors generally provide supervision with different scales and require additional optimization for alignment.

Different from all these methods, our method introduces a prior that can be derived from the depth maps to directly impose constraints on 3D Gaussians in a geometry-aware manner. The coarse-to-fine refining can progressively produce a better prior to constrain Gaussians adaptively, eventually achieving more accurate depth rendering.

\section{Method}
\textbf{Overview. }Our method is illustrated in Fig. \ref{fig:main_pic}.
We aim to learn a set of 3D Gaussians $\{g_j\}_{j=1}^{J}$ from $I$ images $\{v_i\}_{i=1}^{I}$ for high fidelity surface reconstruction. With the learned Gaussians $\{g_j\}$, we can render $\{g_j\}$ into depth maps $\{d'\}$ and fuse them into a TSDF for surface extraction, or RGB images $\{v_i'\}$ for novel view synthesis. 
The key of our method is a self-constrained prior which constrains the learning of 3D Gaussians without data-driven priors for more accurate depth rendering. Our prior does not need an additional learning, and can provide a distance field to impose more specific constraints on 3D Gaussians. For instance, we can remove outlier Gaussians, encourage larger opacity for Gaussians that are close to surfaces or smaller opacity for Gaussians that are far away from surfaces, and move Gaussians nearer to surfaces. Meanwhile, we will regularly update the prior with the most recent rendered depth maps which are more accurate in surface estimation. Additionally, we will shrink the range of area around estimated surfaces each time to progressively impose tightened constraints on the learning of Gaussians.

% \subsection{RGB and Depth Rendering}
% With a set of 3D Gaussians $\{g_j\}$, we can render $\{g_j\}$ into a depth map $d_i$ from a specific view $p_i$ through a splatting procedure $\mathcal{S}$,
% \begin{equation}
% \label{Eq:depth}
% \begin{aligned}
% d_i'=\mathcal{S} (\{g_j\},\{z_{ji}\},p_i),
% \end{aligned} 
% \end{equation}
% \noindent where $z_{ji}$ is the Z coordinate of Gaussian $g_j$ in the camera coordinate system specified by $p_i$. Similarly, we can render $\{g_j\}$ into an RGB image by alpha blending with color $c_j$. i.e., $v_i'=\mathcal{S}(\{g_j\},\{c_j\},p_i)$.

\subsection{Learning Self-Constrained Priors}
\textbf{RGB and Depth Rendering. }With a set of 3D Gaussians $\{g_j\}$, we first render $\{g_j\}$ into a depth map $d_i'$ from a specific view $p_i$ through a splatting procedure $\mathcal{S}$,
% to initialize the Gaussian attributes:
\begin{equation}
\label{Eq:depth}
\begin{aligned}
d_i'=\mathcal{S} (\{g_j\},\{z_{ji}\},p_i),
\end{aligned} 
\end{equation}
\noindent where $z_{ji}$ is the Z coordinate of Gaussian $g_j$ in the 
\begin{wrapfigure}[14]{l}[0em]{0.5\linewidth}
\vspace*{-0.3cm}
\includegraphics[width=\linewidth]{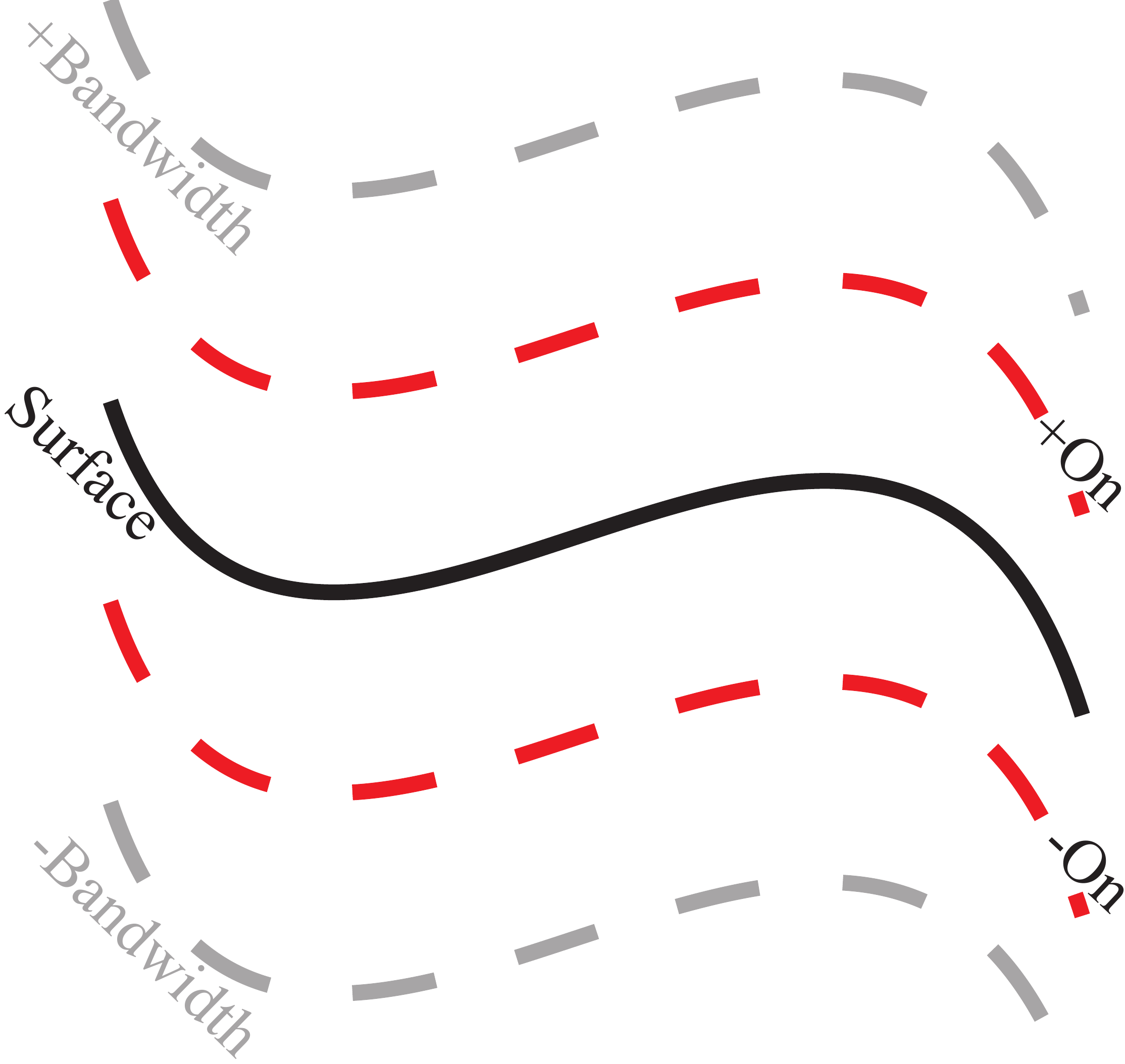} % 设置宽度为半页宽
\vspace{-0.1in}  % 可选，调整图像与上下文的间距
\caption{Illustration of bandwidth and the range of opacity control.}
\label{fig:curves}
\vspace{-0.18in} % 可选，调整图像与下文的间距
\end{wrapfigure}
% \begin{figure}
%     \includegraphics[width=0.5\linewidth]{figs/Curves.pdf} % 设置宽度为半页宽
%     \vspace{-0.1in}  % 可选，调整图像与上下文的间距
%     \caption{Illustration of bandwidth and the range of opacity control.}
%     \label{fig:curves}
%     \vspace{-0.18in} % 可选，调整图像与下文的间距
% \end{figure}
camera coordinate system specified by $p_i$. Similarly, we can render $\{g_j\}$ into an RGB image by alpha blending with color $c_j$. i.e., $v_i'=\mathcal{S}(\{g_j\},\{c_j\},p_i)$.

\noindent\textbf{TSDF Grids. }Our self-constrained prior is a distance field $f^t$ represented as a TSDF grid. It is obtained by fusing depth maps rendered at iteration $t$ with the depth fusion operation $\mathcal{F}$,
\begin{equation}
\label{Eq:fusion}
\begin{aligned}
f^t=\mathcal{F}(\{d_i'(t)\}).
\end{aligned} 
\end{equation}

With the distance field $f^t$, we can regard the zero-level set as a coarse surface estimation, which provides a reference to work with the learning of Gaussians in return. As shown in Fig.~\ref{fig:curves}, beyond the zero-level set, we focus on a narrow band by the surface, which spans a distance field from $-1$ to $1$. We can use a threshold $\sigma^t$ to control the truncation distance at iteration $t$, which determines the width of the narrow band.

% \begin{wrapfigure}{l}{0.5\linewidth}
% \centering
% % \vspace{-0.1in}
% \includegraphics[width=\linewidth]{figs/Curves.pdf}\hspace*{-0.3cm}
% \vspace{-0.15in}
%     \caption{Illustration of bandwidth and the range of opacity control.}
%     \label{fig:curves}
%     % \vspace{-0.18in}
% \end{wrapfigure}

\begin{figure}[t]
    \centering
    \includegraphics[width=\columnwidth]{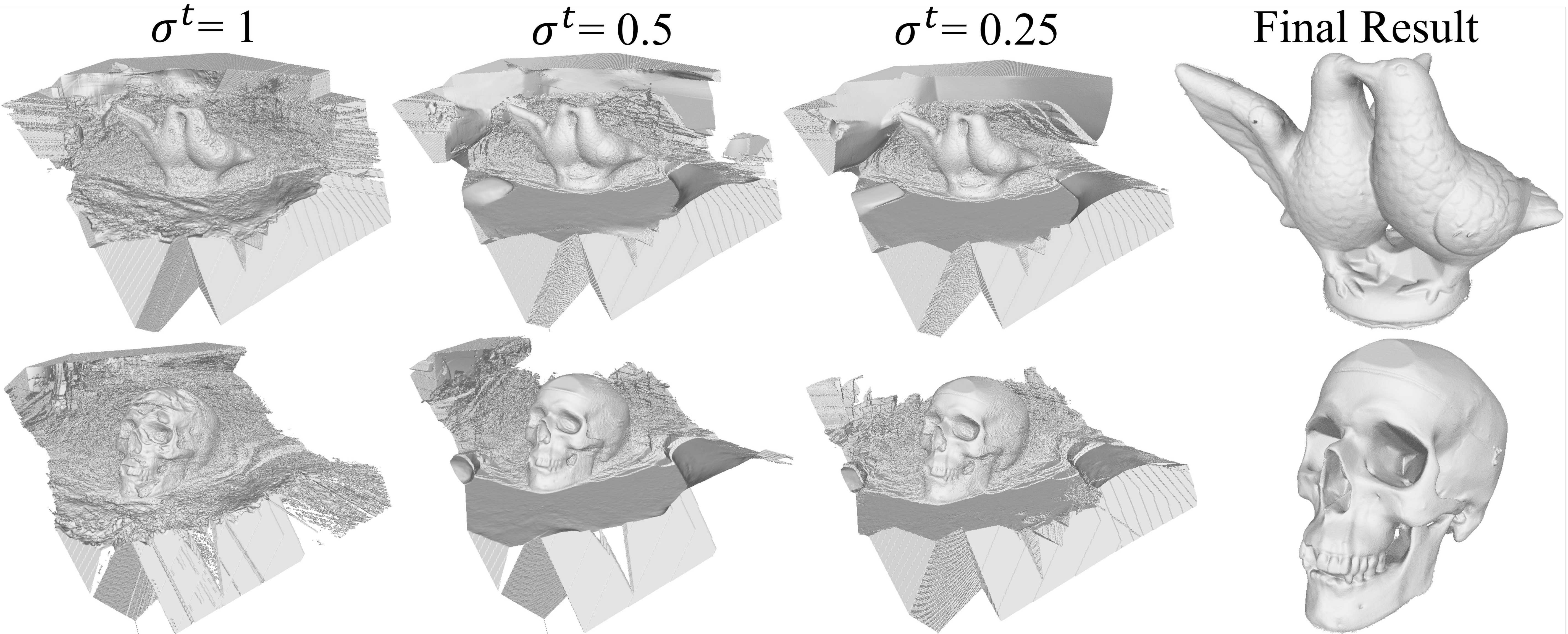}
    \caption{Visualization of periodical update on our prior.}
    \label{fig:period-update}
\end{figure}

\noindent\textbf{Periodic Update. }We update the distance field $f^t$ at regular intervals with the most recent rendered depth maps to improve the coarse surface estimation and balance the additional computation for depth fusion as well. Since the most recent rendered depth maps are more multi-view consistent and accurate, the updated prior can impose more specific constraints on Gaussians and benefit the learning of Gaussians. Moreover, we also progressively reduce the width of the narrow band to strengthen the constraints along with stabilizing the optimization. We show the updated fields $f^t$ with different truncation distances threshold $\sigma^t$ in Fig.~\ref{fig:period-update}.

\begin{figure}[t]
    \centering
    \includegraphics[width=\columnwidth]{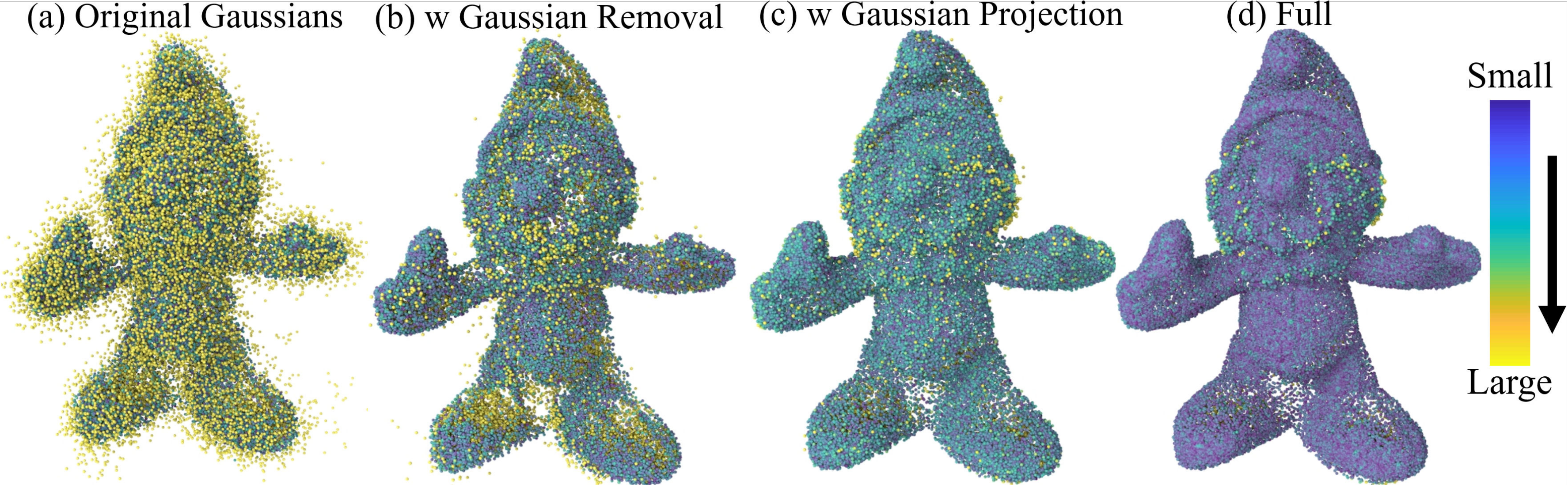}
    \caption{Visualization of Gaussian centers with each constraint. The error map indicates the distance to the ground truth surface.}
    \label{fig:loss constraints}
    \vspace{-0.3Cm}
\end{figure}

% \newpage
\subsection{Constraints with a Self-Constrained Prior}
\noindent\textbf{Removal of Gaussian Outliers. }Once we have a distance field $f^t$, we can remove the Gaussians that are outside the narrow band as specified in Fig.~\ref{fig:curves}. We aim for a more compact Gaussian distribution so that the rendered depth can get less negative impact of Gaussian outliers that are far away from the surface estimation. Specifically, we interpolate the signed distance $s_j$ at the center $\mu_j$ of each Gaussian $g_j$ from the field $f^t$ using the trilinear interpolation, i.e., $s_j=f^t(\mu_j)$. If the interpolated $s_j$ is either $1$ or $-1$, we regard this Gaussian as an outlier and remove it from the Gaussian set. Fig.~\ref{fig:loss constraints} (b) shows Gaussian removal can significantly reduce the impact brought by Gaussian outliers.
% \enlargethispage*{0pt}

\noindent\textbf{Constraints on Opacity. }Although the field $f^t$ remains an estimation, it is adequate to present a coarse surface inference. Unlike the previous methods~\cite{yu2024gaussian,lyu20243dgsr} that directly learn a mapping from SDFs to opacity, 
% \begin{wrapfigure}{l}{0.5\linewidth}
% \centering
% % \vspace{-0.1in}
% \includegraphics[width=\linewidth]{figs/Curves.pdf}\hspace*{-0.3cm}
% \vspace{-0.15in}
%     \caption{Illustration of bandwidth and the range of opacity control.}
%     \label{fig:curves}
%     \vspace{-0.18in}
% \end{wrapfigure}
we encourage opacity to be maximal on the surface estimation and minimal at the band boundary. Thus, we split all Gaussians within the band into two subsets, one is the on-surface subset $\mathcal{N}_{on}$, the other is off-surface subset $\mathcal{N}_{off}$, as illustrated by the range of the two subsets in Fig.~\ref{fig:curves}. We use a signed distance threshold $\delta^t$ to label each Gaussian into one of the two subsets, where we interpolate the $s_j$ of each Gaussian from $f^t$, we have $\mathcal{N}_{on}=\{g_j||s_j|\leq \delta^t\}$, $\mathcal{N}_{off}=\{g_j|\delta^t < |s_j|\leq 1 \}$. Therefore, the constraint on opacity is formulated as,
% \newpage
\begin{equation}
\label{Eq:scploss}
L_{SCP} =
\frac{1}{M}
(\sum_{g_k \in \mathcal{N}_{on}} \varepsilon_k(o_k - 1)^2
+
\sum_{g_{k'} \in \mathcal{N}_{off}} \varepsilon_{k'}o_{k'}^2),
\end{equation}
\noindent where $o$ is the opacity of Gaussian $g$, $M$ is the total number of Gaussians. Both $\varepsilon_k$ and $\varepsilon_{k'}$ are weights for the importance of each sample, according to its interpolated distance to the surface, which is formulated as $\varepsilon_j=1/(1+|s_j|)^2$.\\
% \begin{figure}{l}{0.5\linewidth}
% \centering
% % \vspace{-0.1in}
% \includegraphics[width=\linewidth]{figs/Curves.pdf}
% \vspace{-0.15in}
%     \caption{Illustration of bandwidth and the range of opacity control.}
%     \label{fig:curves}
%     \vspace{-0.15in}
% \end{figure} 
\noindent\textbf{Moving Toward Surfaces. }Within the band, we leverage the signed distance field $f^t$ to move all Gaussians toward the surface. Fig.~\ref{fig:loss constraints} (c) shows that the pulling can move most of Gaussians in Fig.~\ref{fig:loss constraints} (a) closer to surfaces. We leverage the interpolated signed distances and gradients to project all Gaussians onto the surface. Unlike previous methods~\cite{zhang2024gspull,li2025gaussianudf}, we do not involve the pulling into the iterative optimization for stabilizing the optimization. Instead, we regard the projection as an update of Gaussian positions. %update the position of Gaussians after each densification.
For each Gaussian $g_j$, we first interpolate the signed distance $s_j$ at the center $\mu_j$, and calculate its gradient $\nabla f^t(\mu_j)$ using finite difference in the grid-based field $f^t$,

\begin{equation}
\label{Eq:rgbloss}
\begin{aligned}
\nabla f^t(\mu_j) =
\frac{1}{2\epsilon}
\begin{bmatrix}
f^t(x_j+\epsilon, y_j, z_j) - f^t(x_j-\epsilon, y_j, z_j) \\
f^t(x_j, y_j+\epsilon, z_j) - f^t(x_j, y_j-\epsilon, z_j) \\
f^t(x_j, y_j, z_j+\epsilon) - f^t(x_j, y_j, z_j-\epsilon)
\end{bmatrix},
% &\nabla f(q) = \frac{f(q + \epsilon) - f(q - \epsilon)}{2\epsilon},
\end{aligned} 
\end{equation}

\noindent where $\epsilon$ denotes the displacement which gets set to $1\times10^{-4}$, and $\mu_j=[x_j,y_j,z_j]$. With the interpolated signed distance $s_j$ and the gradient $\nabla f^t(\mu_j)$, we update the position by

\begin{equation}
\label{Eq:updatepull}
\begin{aligned}
\mu_j\leftarrow \mu_j-s_j*\nabla f^t(\mu_j).
\end{aligned} 
\end{equation}
\noindent\textbf{Constraint Scheduling. }We schedule these constraints within 3DGS. We impose the opacity constraint in each iteration, and apply pulling and then outlier removal strategies after each densification. With densified Gaussians, we first pull all Gaussians toward the surface using Eq.~\ref{Eq:updatepull}, and then remove the Gaussians outliers that are still outside the band. Fig.~\ref{fig:loss constraints} (d) illustrates the Gaussians with all constraints.

\subsection{Loss Functions}
We use planar Gaussians in 3DGS for better geometry representation. Besides, the constraint on opacity with our self-constrained prior in Eq.~\ref{Eq:scploss}, we also use loss terms to evaluate the RGB rendering errors and multi-view consistency on depth maps. Specifically, we use $L_{RGB}$ to evaluate the error of rendering $v'$ to the input image $v$ with a mean absolute error (MAE), a structural similarity (SSIM), and the multi-view normalized cross correlation (NCC).

Based on that, to learn a consistent surface from multi-view depth maps, we also leverage $L_{Depth}$ to make the per-ray depth distribution thinner and more consistent,

\begin{equation}
\label{Eq:lossnormal}
\begin{aligned}
L_{Depth}=\sum_{u=0}^{U-1}\sum_{u'=0}^{u-1}\omega_u\,\omega_{u'}\,(\rho_u-\rho_{u'})^2,
\end{aligned} 
\end{equation}
where $\{\omega_u,\omega_{u'}\}$ is the alpha-blending weights of the Gaussian, $\rho_u$ is the intersection depth, $U$ is the count of Gaussian on a ray, $u$ and $u'$ index the Gaussians along the ray.

To align Gaussians with actual surface, we introduce a normal regularization for accurate geometry approximation.
% \begin{equation}
% \label{Eq:l_normal_total}
% \begin{aligned}
% L_{Normal} = \lambda_2L_{NS} + \lambda_3L_{NM}.
% \end{aligned}
% \end{equation}
We first employ $L_{NS}$ to evaluate the consistency between the rendered normal $n_i$ and the derived normal $n_i'$ from rendered depth, 
\begin{equation}
\label{Eq:lossnormal}
\begin{aligned}
L_{NS} \;=\; \frac{1}{|I|}\sum_{i\in I} \eta_k \,\bigl\|\, n_i - n_i' \,\bigr\|_{1},
%L_{NS} \;=\; \sum_{k}\omega_{k}\bigl(n_k -n_k'\bigr),
\end{aligned}
\end{equation}
where $\eta_k= (1 - \|\nabla I\|)^2$ serves as per-pixel weights. However, $L_{NS}$ struggles to ensure the alignment of local details and tends to smooth the global shape.

To address this, we incorporate cross-view consistency with the homography matrix $H_{rm}$ to align local geometry between the reference view $r$ and the neighboring view $m$,
\begin{equation}
\label{eq:homography_rm}
H_{rm}
= C_m\!\left(
R_{rm} \;+\; \frac{T_{rm}\, n_r^{\!\top}}{\,\tau_r\,}
\right) C_r^{-1},
\end{equation}
where $\{C_r,C_m\}$ are camera intrinsics, $R_{rm}$ and $T_{rm}$ are relative rotation and translation, $n_r$ is the plane normal in $r$, and $\tau_r$ is the plane distance in reference camera coordinates. Therefore, we constrain the geometry consistency with $L_{NM}$ below,
\begin{equation}
\label{Eq:l_mv}
L_{NM}
= \frac{1}{|V|}\sum_{p_r\in V}
\left\|\,h_r - H_{mr}\,H_{rm}\,h_r\,\right\|_{1},
\end{equation}
where $V$ denotes the set of valid pixels in $r$, $h_r$ is the homogeneous pixel coordinate in $r$, and $\{H_{mr},H_{rm}\}$ are the homography matrices.

\noindent Overall, we minimize the loss function $L$ by,
\begin{equation}
\label{Eq:loss}
\begin{aligned}
L = L_{RGB} + \lambda_1L_{Depth}+ \lambda_2L_{NS} + \lambda_3L_{NM} + \lambda_4L_{SCP},
\end{aligned} 
\end{equation}
where $\{\lambda_1,\lambda_2,\lambda_3,\lambda_4\}$ are the balance weights, and we set them as $\lambda_1=0.01$, $\lambda_2=0.1$, $\lambda_3=0.1$, and $\lambda_4=0.01$.

\section{Experiments}
\subsection{Experiment Setup}
\begin{figure*}[!t]
    \centering
    \includegraphics[width=\textwidth]{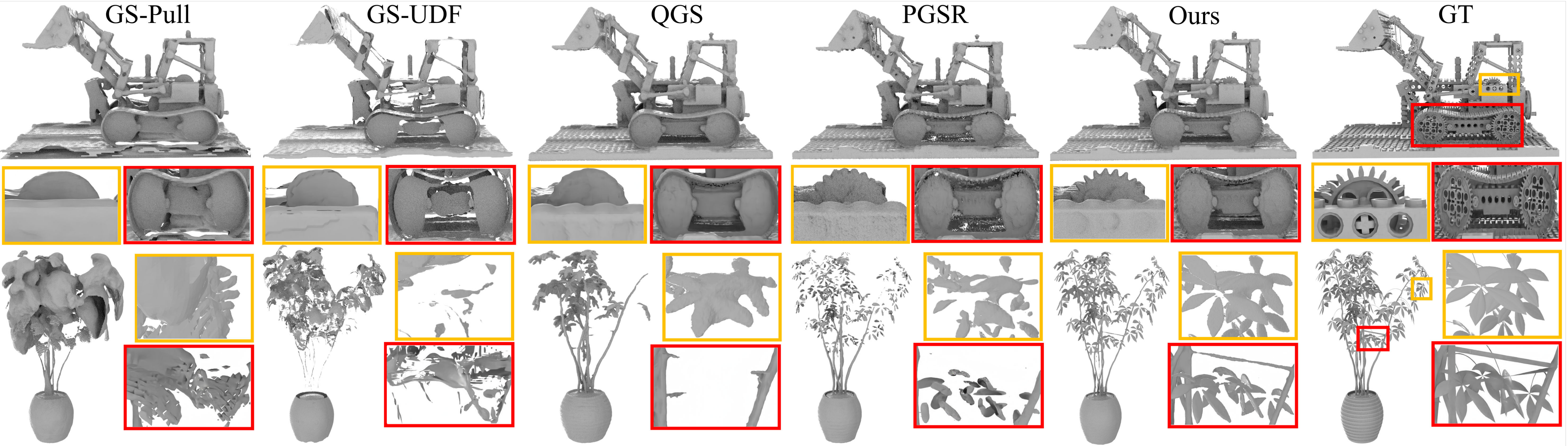}
    \caption{Visual comparison of reconstruction on NeRF-Synthetic dataset.}
    \label{fig:nerf-vis-recon}
\end{figure*}

\textbf{Datasets and Metrics. }We evaluate our method on four datasets with synthetic and real scanned scenes, including: NeRF-Synthetic \cite{mildenhall2020nerf}, DTU \cite{jensen2014large}, Tanks and Temples (TNT) \cite{Knapitsch2017tanks}, and Mip-NeRF 360 \cite{barron2022mip}. For NeRF-Synthetic, we report the L1 Chamfer Distance ($\text{CD}_{L1}$) between the ground-truth point cloud and the predicted surface. Meanwhile, we report PSNR for view accuracy. For DTU, we also analyze the performance in geometry inference and image rendering with CD and PSNR. 
For TNT, we report the F1-score to measure the alignment between the predicted surface and the ground-truth point cloud. For Mip-NeRF 360, we employ PSNR, SSIM, and LPIPS as evaluation metrics for large-scale novel view synthesis. We follow 2DGS to extract surfaces with TSDF-Fusion under the same settings.\\
\textbf{Baselines. }We compare our method with state-of-the-art methods with different learning strategies, including methods with explicit representations \cite{kerbl20233d,huang20242dgs,zhang2024gspull,dai2024high,lyu20243dgsr,yu2024gaussian,yu2024gsdf}, implicit representations \cite{wang2021neus,yariv2021volsdf,mildenhall2020nerf,wang2023neus2,li2023neuralangelo,darmon2022neuralwarp,muller2022instant,yariv2023bakedsdf,barron2022mip}, and various constraints  \cite{chen2024pgsr,dai2024high,li2025gaussianudf,guedon2024sugar,turkulainen2024dnsplatter}. We evaluate the performance in reconstruction and rendering
under the same settings.

\subsection{Results and Evaluation}
\begin{figure}[t]
    \centering
    \includegraphics[width=\columnwidth]{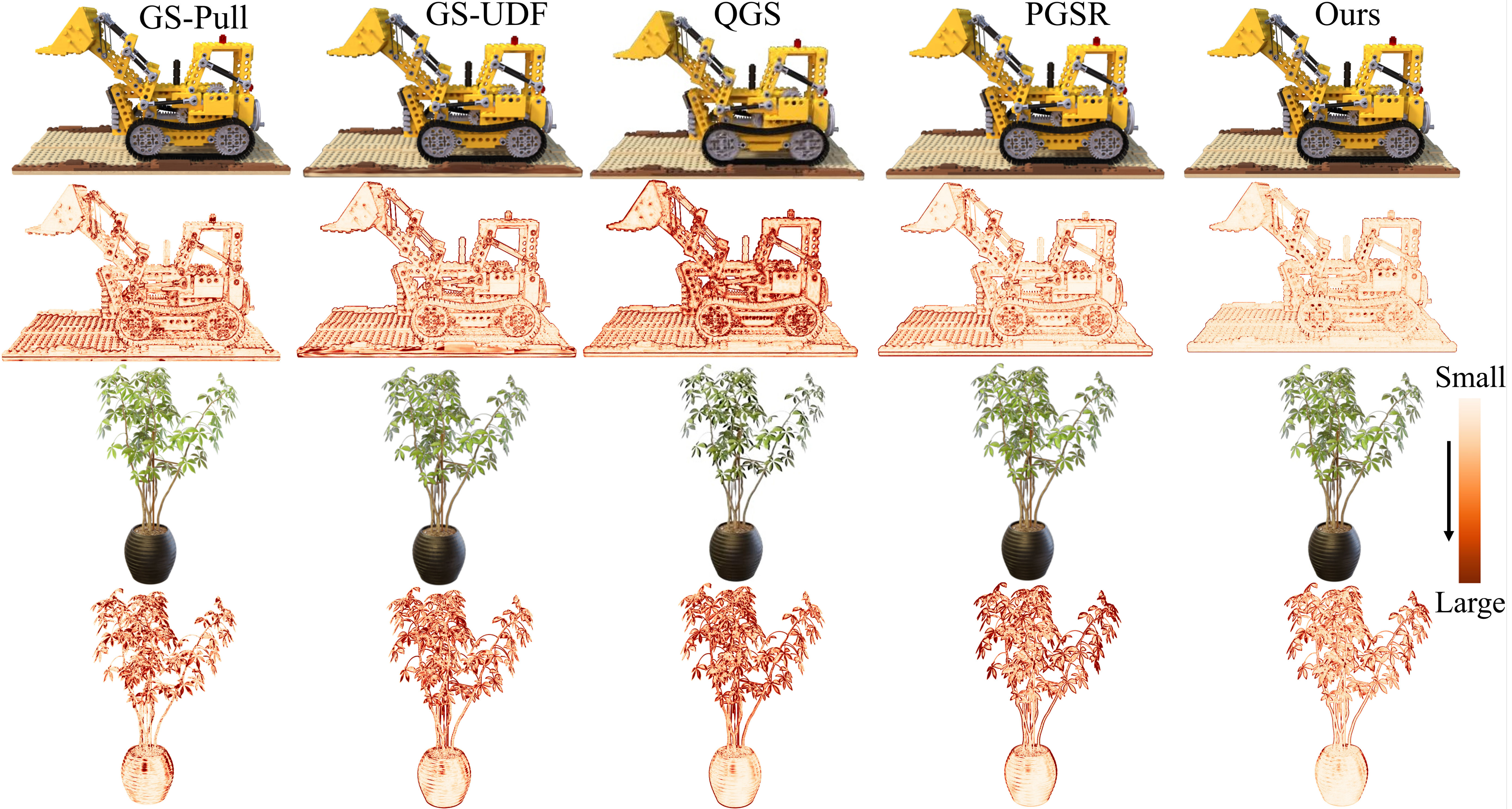}
    \caption{Error map comparison of rendering on NeRF-Synthetic.}
    \label{fig:nerf-vis-render}
    \vspace*{-0.6cm}
\end{figure}

\noindent\textbf{Comparisons on NeRF-Synthetic. }We first evaluate our method on synthetic scenes and present quantitative and visual comparisons in Tab.~\ref{tab:nerf-cd}, Fig.~\ref{fig:nerf-vis-recon}, and Fig.~\ref{fig:nerf-vis-render}, respectively. As reported in Tab.~\ref{tab:nerf-cd}, our method outperforms all baselines in both CD and PSNR metrics. We further present geometry comparison results in Fig.~\ref{fig:nerf-vis-recon}. PGSR~\cite{chen2024pgsr} enforces multi-view consistency but still struggles to recover complete geometric details and complex structures. In contrast, our method leverages depth cues and 3D geometric constraints within the implicit field to 
% align Gaussians with the underlying surfaces and 
achieve more complete geometry. Meanwhile, different from GS-Pull~\cite{zhang2024gspull} and GS-UDF~\cite{li2025gaussianudf} that fit 3D Gaussians with gradients to learn implicit fields, our method extracts stable implicit priors from depth maps for more stable geometry inference. The comparisons of error maps on renderings in Fig.~\ref{fig:nerf-vis-render} also demonstrate that our method produces smaller errors in challenging regions.
\begin{table}
  \centering
  \scriptsize
  \caption{Quantitative comparisons in terms of $\mathrm{CD}_{L1}$ ($\times 100$) and PSNR on the NeRF-Synthetic dataset.}
  \label{tab:nerf-cd}
  \begin{adjustbox}{width=0.8\columnwidth,center}
  \begin{tabular}{c|c|c|c}
    \toprule
    Class & \multicolumn{1}{c|}{Method} & $\mathrm{CD}_{L1}(\times 100) \downarrow$ & PSNR $\uparrow$  \\
    \midrule
    \multirow{2}{*}{Implicit} 
      & NeuS \cite{wang2021neus}       & 2.33 & 30.20 \\
      & NeRO \cite{liu2023nero}    & 1.92  & 27.48 \\
      & VolSDF \cite{yariv2021volsdf}    & 2.86  & 27.96 \\
    \midrule
    \multirow{5}{*}{Explicit} 
      & 2DGS \cite{huang20242dgs}       & 2.26  & 33.07 \\
      & GS-UDF \cite{li2025gaussianudf}    & 2.25   & 33.37 \\
      & GS-Pull \cite{zhang2024gspull}    & 2.31  & 33.29 \\
      & PGSR  \cite{chen2024pgsr}     & 2.18  & 34.05 \\
      & QGS \cite{zhang2025quadratic}    & 2.04  & 30.41 \\
      & Ours & \textbf{1.87} & \textbf{34.21} \\
    \bottomrule
  \end{tabular}
  \end{adjustbox}
\end{table}

\begin{table*}[!t]
  \centering
  \caption{Quantitative comparisons in terms of CD on the DTU dataset. }
  \label{tab:dtu-cd}
  \begin{adjustbox}{width=\textwidth,center}
  \begin{tabular}{c|c|ccccccccccccccc|cc}
    \toprule
     Class & Methods & 24 & 37 & 40 & 55 & 63 & 65 & 69 & 83 & 97 & 105 & 106 & 110 & 114 & 118 & 122 & Mean &Time  \\
    \midrule
    \multirow{5}{*}{\makecell[c]{Implicit}}
 & NeRF ~\cite{mildenhall2020nerf}       & 1.90 & 1.60 & 1.85 & 0.58 & 2.28 & 1.27 & 1.47 & 1.67 & 2.05 & 1.07 & 0.88 & 2.53 & 1.06 & 1.15 & 0.96 & 1.49 &$>$12h \\
 & VolSDF ~\cite{yariv2021volsdf}        & 1.14 & 1.26 & 0.81 & 0.49 & 1.25 & 0.70 & 0.72 & 1.29 & 1.18 & 0.70 & 0.66 & 1.08 & 0.42 & 0.61 & 0.55 & 0.86 &$>$12h \\
 & NeuS ~\cite{wang2021neus}              & 1.00 & 1.37 & 0.93 & 0.43 & 1.10 & 0.65 & 0.57 & 1.48 & 1.09 & 0.83 & 0.52 & 1.20 & 0.35 & 0.49 & 0.54 & 0.84 &$>$12h \\
 & Neuralangelo ~\cite{li2023neuralangelo} & 0.37 & 0.72 & 0.35 & 0.35 & 0.87 & 0.54 & 0.53 & 1.29 & 0.97 & 0.73 & 0.47 & 0.74 & 0.32 & 0.41 & 0.43 & 0.61 &$>$12h\\
 & NeuralWarp ~\cite{darmon2022neuralwarp}   & 0.49 & 0.71 & 0.38 & 0.38 & 0.79 & 0.82 & 0.82 & 1.20 & 1.06 & 0.68 & 0.66 & 0.74 & 0.41 & 0.63 & 0.51 & 0.68 &$>$10h \\
 \midrule
     \multirow{9}{*}{\makecell[c]{Explicit}}
 & 2DGS~\cite{huang20242dgs}         & 0.48 & 0.91 & 0.39 & 0.39 & 1.01 & 0.83 & 0.81 & 1.36 & 1.27 & 0.76 & 0.70 & 1.40 & 0.40 & 0.76 & 0.52 & 0.80 &\textbf{19.2min} \\
 & GS-Pull~\cite{zhang2024gspull}      & 0.51 & 0.56 & 0.46 & 0.39 & 0.82 & 0.67 & 0.85 & 1.37 & 1.25 & 0.73 & 0.54 & 1.39 & 0.35 & 0.88 & 0.42 & 0.75 &22min \\
 & GS-UDF~\cite{li2025gaussianudf}       & 0.62 & 0.67 & 0.43 & 0.42 & 0.83 & 0.86 & 0.72 & 1.20 & 1.03 & 0.68 & 0.61 & 0.63 & 0.43 & 0.56 & 0.52 & 0.68 &25min \\
  % & 3DGSR~\cite{lyu20243dgsr}        & 0.44 & 0.96 & 0.40 & 0.36 & 1.02 & 0.80 & 0.64 & 1.20 & 1.08 & 0.97 & 0.54 & 0.72 & 0.37 & 0.52 & 0.42 & 0.70 \\
  & SuGaR~\cite{guedon2024sugar}        & 1.47 &1.33 &1.13 &0.61 &2.25 &1.71 &1.15 &1.63 &1.62 &1.07 &0.79 &2.45 &0.98 &0.88 &0.79 &1.33 &1h\\
 & GOF~\cite{yu2024gaussian}          & 0.50 & 0.82 & 0.37 & 0.37 & 1.12 & 0.74 & 0.73 & 1.18 & 1.29 & 0.68 & 0.77 & 0.90 & 0.42 & 0.66 & 0.49 & 0.74 &1h \\
  & PGSR~\cite{chen2024pgsr}         & 0.34 & 0.58 & \textbf{0.29} & \textbf{0.29} & 0.78 & 0.58 & 0.54 & \textbf{1.01} & 0.73 & \textbf{0.51} & 0.49 & 0.69 & 0.30 & \textbf{0.35} & 0.38 & 0.53 &40min \\
 & GSDF~\cite{yu2024gsdf}         & 0.58 & 0.93 & 0.46 & 0.37 & 1.30 & 0.77 & 0.73 & 1.58 & 1.28 & 0.76 & 0.58 & 1.22 & 0.37 & 0.51 & 0.51 & 0.80 &32min \\
  & QGS~\cite{zhang2025quadratic}       & 0.38 & 0.62 & 0.37 & 0.38 & 0.75 & 0.55 & 0.51 & 1.12 & \textbf{0.68} & 0.61 & 0.46 & \textbf{0.58} & 0.35 & 0.41 & 0.40 & 0.54 &48min \\
 & Ours         & \textbf{0.32} & \textbf{0.55} & 0.30 & 0.31 & \textbf{0.74} & \textbf{0.54} & \textbf{0.48} & 1.06 & 0.69 & 0.57 & \textbf{0.45} & \textbf{0.58} & \textbf{0.29} & \textbf{0.35} & \textbf{0.33} & \textbf{0.50} &42min \\
    \bottomrule
  \end{tabular}
  \end{adjustbox}
\end{table*}
\begin{figure*}
    \centering
    \includegraphics[width=\textwidth]{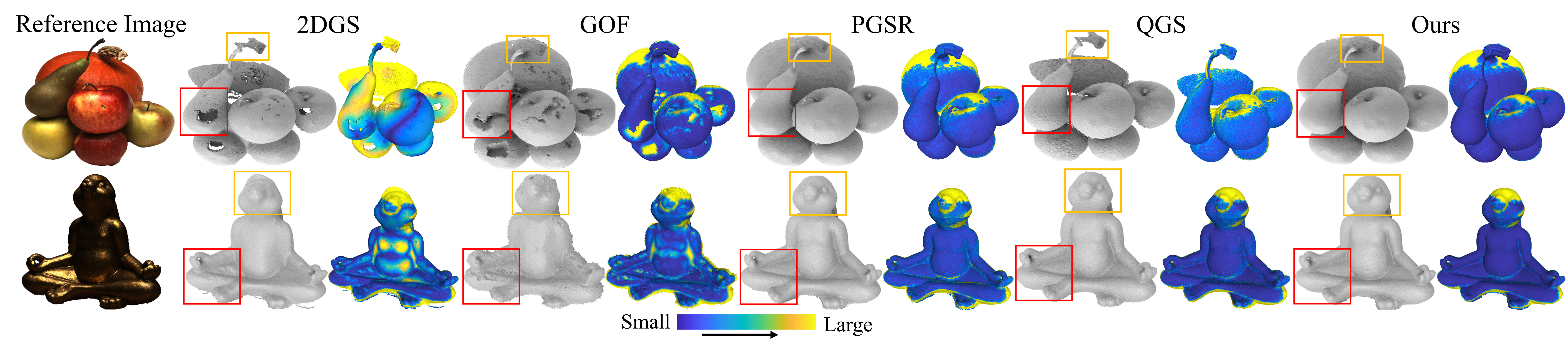}
    \caption{Visual comparison of reconstruction on DTU dataset. The error map indicates the distance to the ground truth surface.}
    \label{fig:dtu-vis-recon}
\end{figure*}

\begin{table}[t]
\centering
\caption{Quantitative comparisons in terms of F1-Score on the TNT dataset.}
\label{tab:tnt-f1}
\begin{adjustbox}{width=\columnwidth,center}
\begin{tabular}{c|c|cccccc|c}
\toprule
Class & Methods & Barn & Caterpillar & Courthouse & Ignatius & Meetingroom & Truck & Mean \\
\midrule
\multirow{4}{*}{\makecell[c]{\rotatebox{90}{Implicit}}}
  & NeuS~\cite{wang2021neus}        & 0.29 & 0.29 & 0.17 & 0.83 & 0.24 & 0.45 & 0.38 \\
  & Geo-NeuS~\cite{fu2022geoneus}    & 0.33 & 0.26 & 0.12 & 0.72 & 0.20 & 0.45 & 0.35 \\
  & Neuralangelo~\cite{li2023neuralangelo} &0.70 &0.36 &0.28 &0.89 &0.32 &0.48 &0.50 \\
  & MonoSDF~\cite{yu2022monosdf}     &0.49 &0.31 &0.12 &0.78 &0.23 &0.42 &0.39 \\
 \midrule
\multirow{8}{*}{\makecell[c]{\rotatebox{90}{Explicit}}}
  & 3DGS~\cite{kerbl20233d}        & 0.13 & 0.08 & 0.09 & 0.04 & 0.01 & 0.19 & 0.09 \\
  & SuGaR~\cite{guedon2024sugar}       & 0.14 & 0.16 & 0.08 & 0.33 & 0.15 & 0.26 & 0.19 \\
  & GSurfels~\cite{dai2024high}    & 0.24 & 0.22 & 0.07 & 0.39 & 0.12 & 0.24 & 0.21 \\
  % & 2DGS~\cite{huang20242dgs}        & 0.41 & 0.23 & 0.16 & 0.51 & 0.17 & 0.45 & 0.32 \\
  & 2DGS~\cite{huang20242dgs}        & 0.36 & 0.23 & 0.13 & 0.44 & 0.16 & 0.26 & 0.30 \\
  & PGSR~\cite{chen2024pgsr}        & 0.66 & 0.41 & \textbf{0.21} & 0.80 & 0.29 & 0.60 & 0.50 \\
  & GOF~\cite{yu2024gaussian}         & 0.51 & 0.41 & 0.28 & 0.68 & 0.28 & 0.59 & 0.46 \\
  & GS-Pull~\cite{zhang2024gspull}     & 0.60 & 0.37 & 0.16 & 0.71 & 0.22 & 0.52 & 0.43 \\ 
  & QGS~\cite{zhang2025quadratic}     & 0.55 & 0.40 & 0.28 & \textbf{0.81} & \textbf{0.31} & \textbf{0.64} & 0.50 \\ 
  & Ours        &\textbf{0.66} &\textbf{0.43} &0.20 &0.80 &\textbf{0.31} &\textbf{0.64} &\textbf{0.51} \\
\bottomrule
\end{tabular}
\end{adjustbox}
\vspace{-0.1in}
\end{table}

\begin{figure*}[t]
    \centering
    \includegraphics[width=\textwidth]{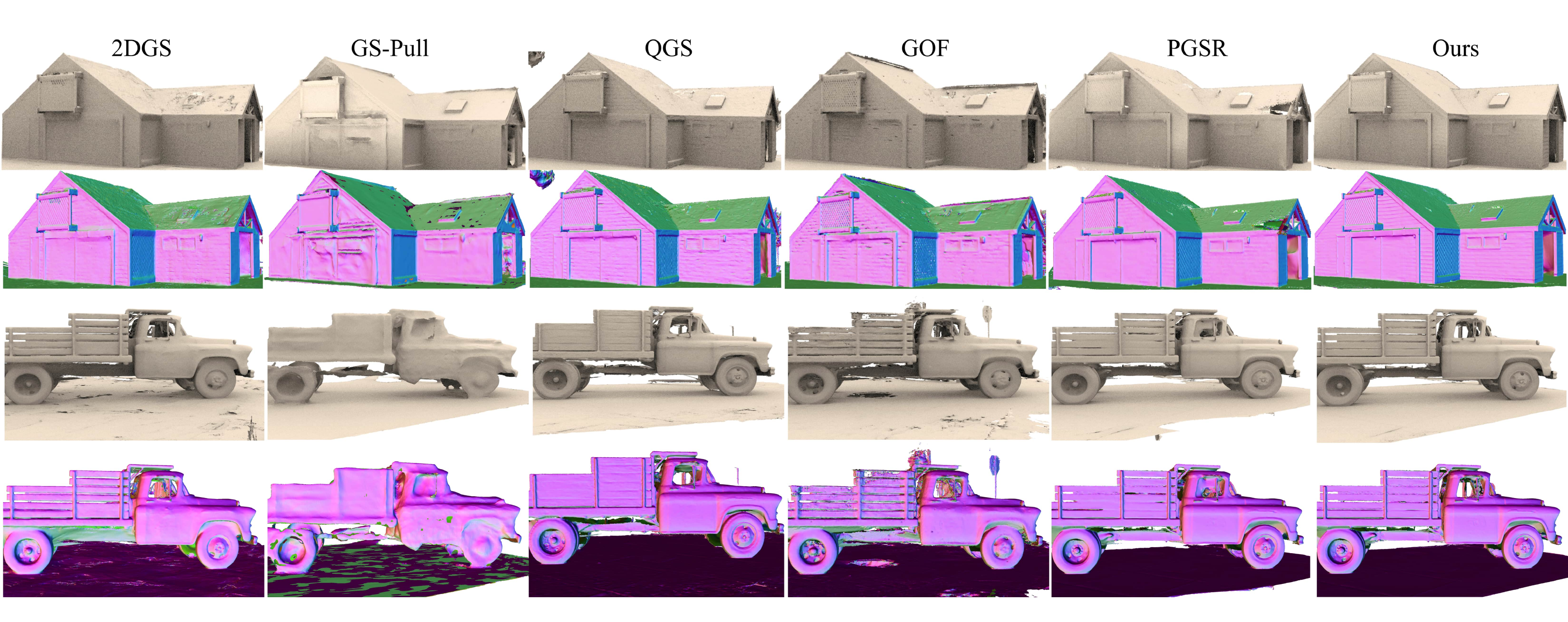}
     \vspace{-0.2in}
    \caption{Visual comparison of reconstruction on TNT dataset, the color indicates the normal direction.}
    \label{fig:tnt-vis-recon}

\end{figure*}

\begin{figure*}
    \centering
    \includegraphics[width=\textwidth]{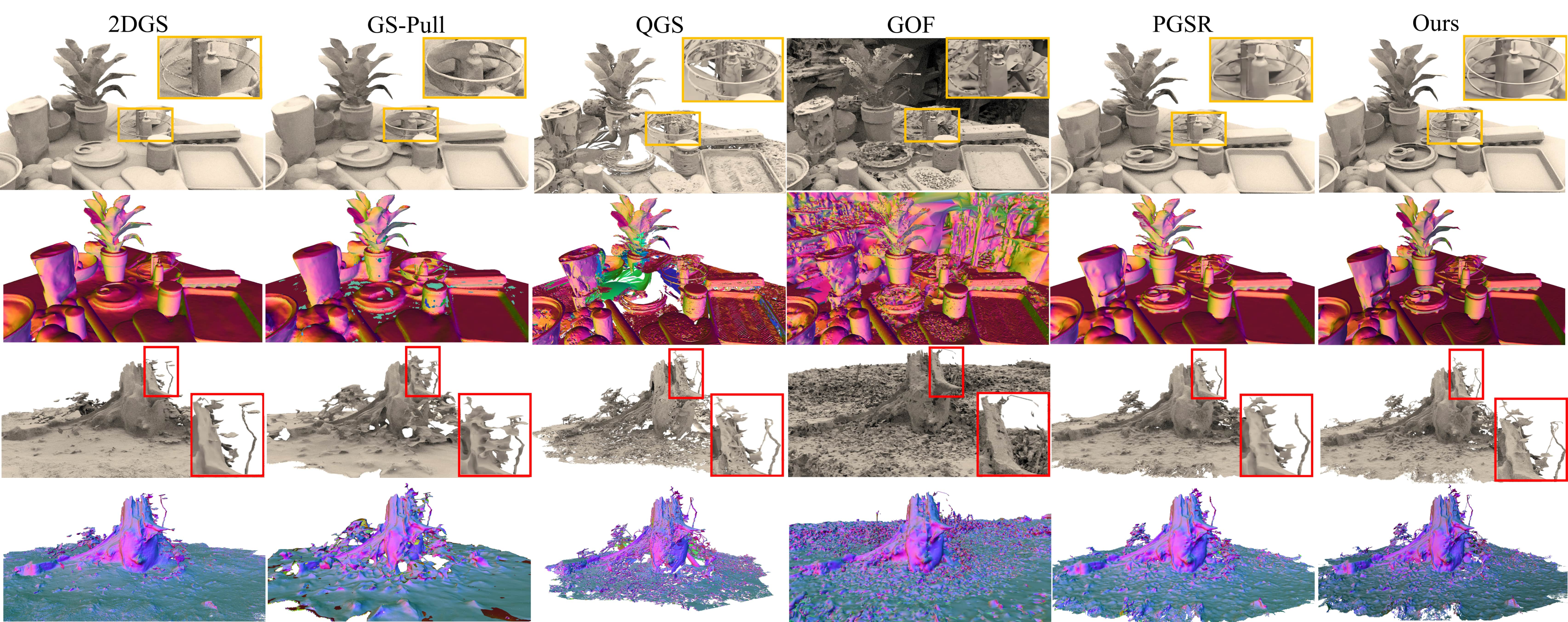}
    \caption{Visual comparison of reconstruction on Mip-NerF 360 dataset, the color indicates the normal direction.}
    \label{fig:mip360-vis-recon}
    \vspace{-0.2in}
\end{figure*}
\begin{table}
  \centering
  \caption{Quantitative comparisons in terms of PSNR, SSIM and LPIPS on the Mip-NeRF 360 dataset.}
  \label{tab:mipnerf360}
  \begin{adjustbox}{width=\linewidth,center}
  \begin{tabular}{c|c|ccc|ccc}
    \toprule
    \multirow{2}{*}{Class} & \multirow{2}{*}{Method} 
    & \multicolumn{3}{c|}{Outdoor Scenes} 
    & \multicolumn{3}{c}{Indoor Scenes} \\
     & & PSNR$\uparrow$ & SSIM$\uparrow$ & LPIPS$\downarrow$ 
       & PSNR$\uparrow$ & SSIM$\uparrow$ & LPIPS$\downarrow$ \\
    \midrule
    \multirow{4}{*}{\rotatebox{90}{Implicit}}
     & NeRF~\cite{mildenhall2020nerf}         & 21.46 & 0.458 & 0.515 & 26.84 & 0.790 & 0.370 \\
     & Ins-NGP~\cite{muller2022instant}  & 22.90 & 0.566 & 0.371 & 29.15 & 0.880 & 0.216 \\
     & BakedSDF~\cite{yariv2023bakedsdf}     & 22.47 & 0.585 & 0.349 & 27.06 & 0.836 & 0.258 \\
     & Mip-NeRF 360~\cite{barron2022mip} & 24.47 & 0.691 & 0.283 & 31.72 & 0.917 & 0.180 \\
    \midrule
    \multirow{7}{*}{\rotatebox{90}{Explicit}}
     & 3DGS~\cite{kerbl20233d}         & 24.64 & 0.731 & 0.234 & 30.41 & 0.920 & 0.189 \\
     & SuGaR~\cite{guedon2024sugar}        & 22.93 & 0.629 & 0.356 & 29.43 & 0.906 & 0.225 \\
     & 2DGS~\cite{huang20242dgs}         & 24.34 & 0.717 & 0.246 & 30.40 & 0.916 & 0.195 \\
     & GS-Pull~\cite{zhang2024gspull}      & 23.76 & 0.703 & 0.278 & 30.78 & 0.925 & 0.182 \\
     & GOF~\cite{yu2024gaussian}          & 24.82 & 0.750 & 0.202 & \textbf{30.79} & 0.924 & 0.184 \\
     & PGSR~\cite{chen2024pgsr}         & 24.45 & 0.730 & 0.224 & 30.41 & 0.930 & 0.161 \\
     & QGS~\cite{zhang2025quadratic}         & 24.32 & 0.706 & 0.242 & 30.45 & 0.919 & 0.184 \\
     & Ours         & \textbf{24.81} & \textbf{0.754} & \textbf{0.200} & 30.56 & \textbf{0.933}  & \textbf{0.155}  \\
    \bottomrule
  \end{tabular}
  \vspace{-0.5in}
  \end{adjustbox}
\end{table}

\noindent\textbf{Comparisons on DTU. }We also validate our method on the real-scanned dataset compared with state-of-the-art approaches. As shown in Tab.~\ref{tab:dtu-cd}, our method achieves the best results across scenes. Compared with implicit methods, our method does not need to learn SDF or priors, which balances both accuracy and efficiency. 2DGS~\cite{huang20242dgs} and QGS~\cite{zhang2025quadratic} apply 2D normal regularization to smooth surfaces. However, 2D priors are insufficient for accurate surface reconstruction, leading artifacts in challenging regions. GOF~\cite{yu2024gaussian} combines Gaussians with opacity fields to improve performance, but constrained by complex opacity modeling. As shown in Fig.~\ref{fig:dtu-vis-recon}, PGSR~\cite{chen2024pgsr} further integrates multi-view consistency constraints to infer more accurate surfaces, but it still struggles to recover local details. Unlike these methods, our method extracts geometric cues directly from 2D depth maps and produces more detailed and continuous surfaces.\\
\noindent\textbf{Comparisons on TNT. }We evaluate the robustness of our method on large-scale scenes in Tanks and Temples (TNT) dataset. Tab.~\ref{tab:tnt-f1} shows that our method achieves the best reconstruction performance among all baselines. GS-Pull~\cite{zhang2024gspull} relies on Gaussian positions for geometry inference, it is still hard to maintain all Gaussians on surfaces when learning large-scale implicit representations. Similarly, GOF~\cite{yu2024gaussian} employs tetrahedral grids to infer complete surfaces. However, the opacity modeling limits grid optimization in large-scale dataset, leading to uneven surfaces and coarse local details. In contrast, our method reconstructs complete surfaces under complex topology using only depth cues and self-supervised geometric constraints, which effectively detect artifacts and improves efficiency. We further provide visual comparisons in Fig.\ref{fig:tnt-vis-recon}. QGS~\cite{zhang2025quadratic} and GS-Pull~\cite{zhang2024gspull} constrain Gaussians to lie on the surface with regularization terms and implicit field gradients, respectively. However, QGS produces artifacts due to inaccurate surface alignment, while GS-Pull loses local details and exhibits normal errors caused by gradient noise. In contrast, our method leverages a stable implicit field to guide Gaussians toward the surface and produces more continuous and smoother surfaces.\\
\noindent\textbf{Comparisons on Mip-NeRF 360. }We further evaluate our method on the Mip-NeRF 360 dataset to validate the performance in novel view synthesis. Tab.~\ref{tab:mipnerf360} presents quantitative comparisons with both explicit and implicit methods. Our method outperforms all baselines in most metrics. Although surfel based or geometry constraints based methods perform well in geometry inference, the rendering performance is worse than 3DGS due to the impact of strict geometry constraints on Gaussian movement. As shown in Fig.~\ref{fig:mip360-vis-recon}, GOF\cite{yu2024gaussian} and GS-Pull~\cite{zhang2024gspull} achieve high completeness on surfaces but struggle to recover local details in indoor scenes. PGSR~\cite{chen2024pgsr} and QGS~\cite{zhang2025quadratic} suffer from local truncations and artifacts due to truncated distance fields, especially in complex outdoor scenes where multi-view supervision alone struggle to maintain surface continuity on textured regions. In contrast, our method aligns Gaussians near the depth rendered surfaces to preserve the rendering fidelity and reconstruction details.

% \clearpage
% \begin{table*}[t]
%   \centering
%   \caption{Ablation studies on DTU dataset.}
%   \vspace{-0.0cm}
%   \label{tab:ablation-study}
%   \resizebox{1.0\textwidth}{!}{%
%   \begin{tabular}{c|cc|ccc|cc|cc|ccc|c}
%     \toprule
%       & \multicolumn{2}{c|}{Self-constrained Priors}
%       & \multicolumn{3}{c|}{Prior Update}
%       & \multicolumn{2}{c|}{Constraints on Gaussians}
%       & \multicolumn{2}{c|}{Opacity Constraints}
%       & \multicolumn{3}{c}{Bandwidth Sizes} \\
%     \cmidrule(lr){2-3}\cmidrule(lr){4-6}\cmidrule(lr){7-8}\cmidrule(lr){9-10}\cmidrule(lr){11-13}
%     \multicolumn{1}{c|}{Methods}
%       & w $f^t$ & w/o $f^t$
%       & $\{f^{0}\}$ & $\{f^{0},f^{1}\}$ & $\{f^{0},f^{1},f^{2}\}$
%       & w/o $f_{remove}^t$ & w/o $f_{proj}^t$
%       & w $L_{SCP}$ & w/o $L_{SCP}$
%       & 1 & 0.5 & 0.25
%       % & \multicolumn{1}{c}{Full} \\
%     \midrule
%     CD$\downarrow$
%       &0.50  &0.56  &0.54  &0.52  &0.50  &0.53  &0.51  &0.50  &0.53  &0.54  &0.53  &0.51  &0.50  \\
%     \bottomrule
%   \end{tabular}}
%   \vspace{-0.2in}
% \end{table*}

\begin{table*}[t]
  \centering
  \caption{Ablation studies on DTU dataset.}
  \label{tab:ablation-study}
  \resizebox{1.0\textwidth}{!}{%
  \begin{tabular}{c|cc|cc|ccc|ccc|cccc}
    \toprule
      & \multicolumn{2}{c|}{Self-constrained Priors}
      & \multicolumn{2}{c|}{Opacity Constraints}
      & \multicolumn{3}{c|}{Supervision Refinement}
      & \multicolumn{3}{c|}{Constraints on Gaussians}
      & \multicolumn{4}{c}{Bandwidth Scales} \\
    \cmidrule(lr){2-3}\cmidrule(lr){4-5}\cmidrule(lr){6-8}\cmidrule(lr){9-11}\cmidrule(lr){12-15}
    \multicolumn{1}{c|}{Methods}
      & w $f^{t}$ & w/o $f^{t}$
      & w $L_{SCP}$ & w/o $L_{SCP}$
      & $\{f^{t}_{0}\}$ & $\{f^{t}_{0},f^{t}_{1}\}$ & $\{f^{t}_{0},f^{t}_{1},f^{t}_{2}\}$
      & w/o $f^{t}_{remove}$ & w/o $f^{t}_{proj}$ & Full
      & 1 & 0.5 & 0.25 & Full \\
    \midrule
    CD$\downarrow$
      &\textbf{0.50} &0.56
      &\textbf{0.50} &0.53
      &0.54 &0.52 &\textbf{0.50}
      &0.53 &0.51 &\textbf{0.50}
      &0.54 &0.53 &0.51 &\textbf{0.50} \\
    \bottomrule
  \end{tabular}}
\end{table*}

\section{Ablation Studies}
To validate the effectiveness of each module on DTU, we evaluate quantitative and visual analyses in Tab.~\ref{tab:ablation-study}.

\begin{figure}
    \centering
    \includegraphics[width=\columnwidth]{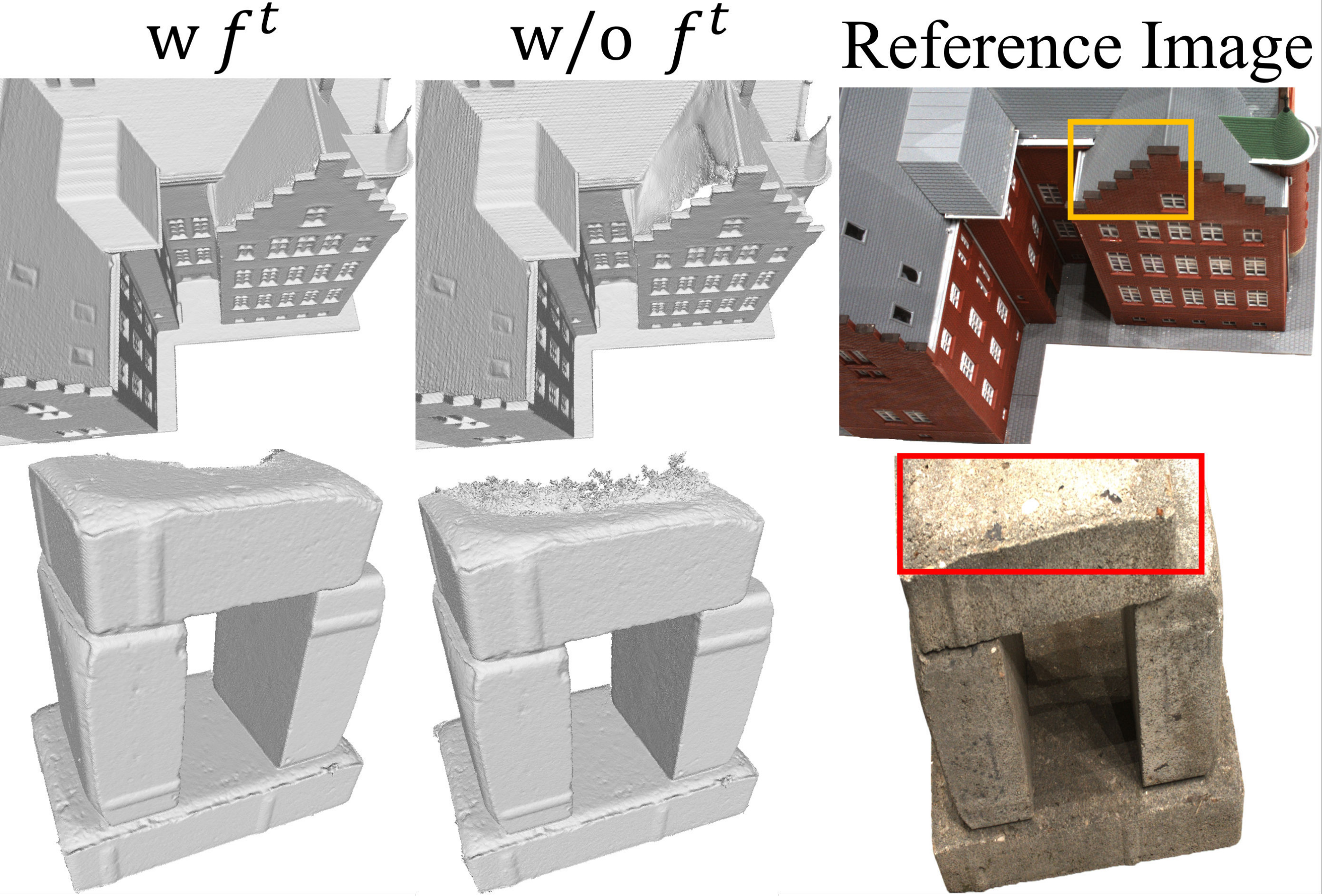}
    \caption{Effect of the self-constrained prior.}
    \label{fig:ab_implicit_supervision}
\vspace{-0.1in}
\end{figure}

\noindent\textbf{Self-constrained Priors. }We first evaluate the effect of the prior $f^{t}$. When we remove $f^{t}$ (denoted as w/o $f^{t}$), we lose the reference to impose constraints. As shown in Tab.~\ref{tab:ablation-study} (``Self-constrained Priors''), the lack of $f^{t}$ leads to a noticeable increase in CD. Visual results in Fig.~\ref{fig:ab_implicit_supervision} further demonstrate that learning Gaussian representations with $f^{t}$ produces more continuous and accurate surfaces.

\begin{figure}
    \centering
    \includegraphics[width=\columnwidth]{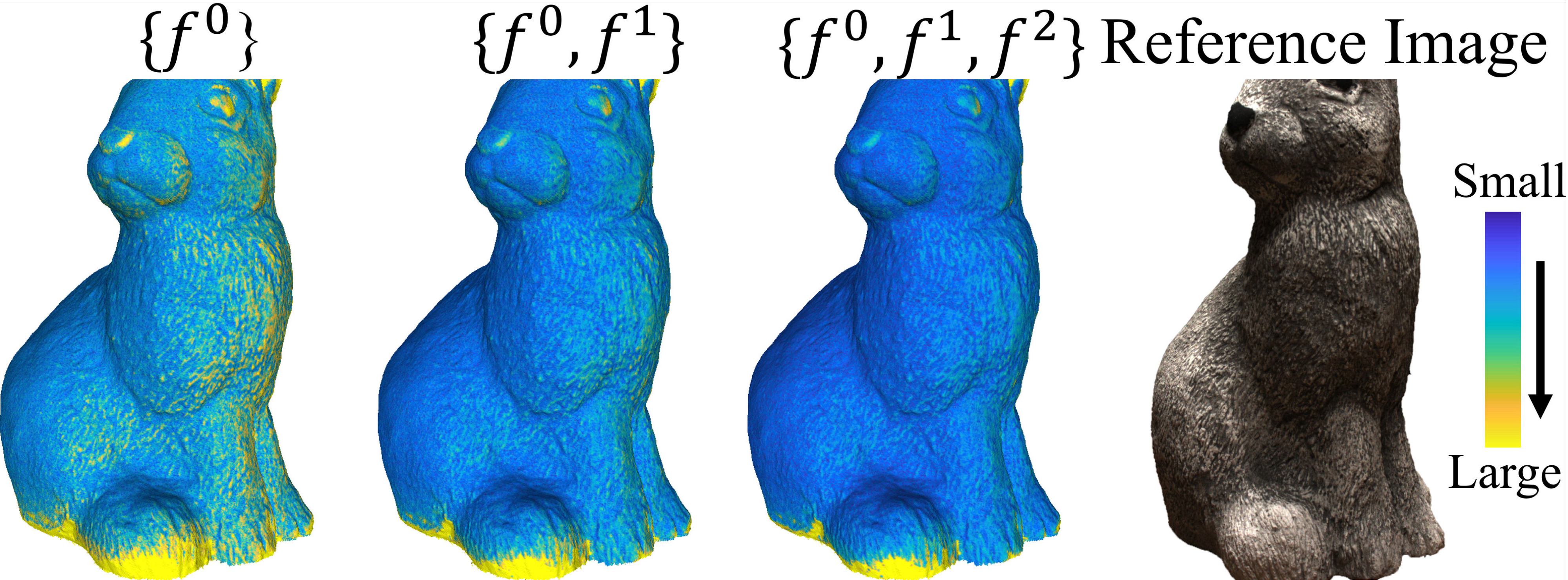}
    \caption{Effect of periodical update on our prior.}
    \label{fig:ab_supervision_refinement}

\end{figure}

\noindent\textbf{Prior Update. }We also access the effect of prior update on reconstruction performance. We set $f^{0}$ as baseline, which is the initial prior. Meanwhile, we update $f^{t}$ from $f^{0}$ to $f^{2}$ for evaluation. As shown in Tab.~\ref{tab:ablation-study} (``Prior Update''), CD error decreases as $f^{t}$ is progressively updated. Furthermore, Fig.~\ref{fig:ab_supervision_refinement} shows that $f^{t}$ guides the Gaussians to recover more continuous and detailed geometry with periodic refinement.

\begin{figure}
    \centering
    \includegraphics[width=\columnwidth]{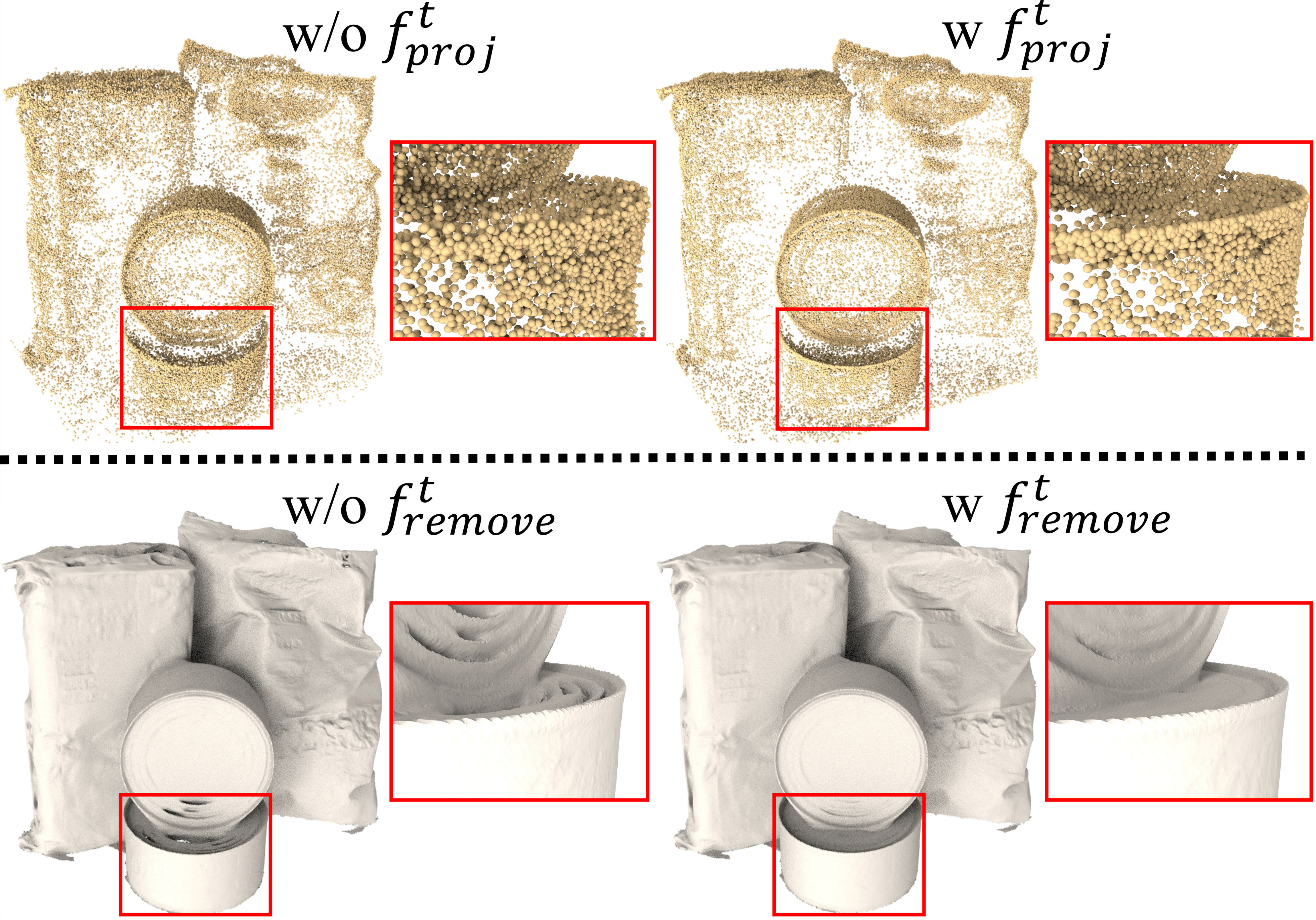}
    \caption{Effect of Gaussian Removal and Projection.}
    \vspace{-0.2in}
    \label{fig:ab_gaussian_update}
\end{figure}

\noindent\textbf{Constraints on Gaussians. }We further analyze the effect of Gaussian constraints. As shown in Tab.~\ref{tab:ablation-study} (``Constraints on Gaussians''), we first remove the Gaussian projection (w/o $f^{t}_{proj}$), which leads CD to increase a bit. We further disable the Gaussian removal ($f^{t}_{remove}$) and the performance gets worse. As shown in Fig.~\ref{fig:loss constraints} and Fig.~\ref{fig:ab_gaussian_update}, the 3D Gaussians struggle to move toward the surfaces without $f^{t}_{proj}$, while $f^{t}_{remove}$ helps 3D Gaussians produce continuous surfaces and removes artifacts beyond the bandwidth.

        % \vspace{-0.2in}

\noindent \textbf{Opacity Constraints. }To validate the effect of the opacity arrangement term $L_{SCP}$, we remove it (denoted as w/o $L_{SCP}$) and optimize the Gaussian opacities only with 2D supervisions. As shown in Tab. ~\ref{tab:ablation-study} ("Opacity Constraints"), removing $L_{SCP}$ leads to degraded reconstruction. Although learning with multi-view constraints can recover the global shapes, but 2D constraints struggle to capture local details with complex geometry.

\noindent\textbf{Bandwidth Sizes. }To evaluate the effect of the bandwidth shrinking strategy, we compare a fixed bandwidth with progressively decreased thresholds $\sigma^{t} = {1, 0.5, 0.25}$. As shown in Fig.~\ref{fig:period-update}, a narrower bandwidth allows $f^{t}$ to capture more accurate geometric distributions and also impose more compact constraints on Gaussians. Tab.~\ref{tab:ablation-study} further shows that the CD error decreases as the bandwidth becomes tighter, and the threshold reduction from 1 to 0.25 (Full) achieves the best performance.

\section{Conclusion}
We present self-constrained priors to recover high fidelity surfaces with 3DGS. We show that a self-constrained prior can be easily obtained by fusing rendered depth images without other information. With the prior, we can effectively impose constraints on specific Gaussians including the removal of outlier Gaussians, opacity control, and move Gaussians closer to the surface. Moreover, periodically update the prior also helps stabilize the optimization and progressively approach convergence. Our evaluations justified our design and show advantages over the latest methods.
\section{Acknowledgment}
This work was partially supported by Deep Earth Probe and Mineral Resources Exploration ­—  National Science and Technology Major Project (2024ZD1003405), and the National Natural Science Foundation of China (62272263, W2542037).

{
    \small
    \bibliographystyle{ieeenat_fullname}
    \bibliography{main}
}

\clearpage

\newpage

\setcounter{page}{1}
\maketitlesupplementary
\setcounter{section}{0}

\section{Implementation Details}
We follow 3DGS~\cite{kerbl20233d} in the experimental setting. We adopt the same optimization schedule to train 30000 iterations with Adam optimizer and start from randomly initialized Gaussian attributes. The TSDF grid $f^{t}$ is updated every 5000 iterations up to iteration 20000. The bandwidth scaling is set to $1, 0.5, 0.25$ in each updated $f^t$, respectively. We enable $L_{SCP}$ at 10000 iteration and use a decision threshold $\delta^{t}=0.3$ on $f^{t}$ to label Gaussians for $L_{SCP}$. All experiments run on a single NVIDIA RTX 4090 GPU.\\
\section{Optimization Details}
\textbf{Rendering Losses. }During rendering optimization, we supervise the learning of 3D Gaussians using a unified image reconstruction loss $L_{RGB}$, which comprises three components: a MAE loss $L_{MAE}$ that constrains pixel-level differences, a SSIM loss $L_{SSIM}$ that preserves structural consistency, and a NCC loss $L_{NCC}$ that enforces multi-view consistency. Together, these terms enable the model to achieve accurate color reconstruction, maintain structural fidelity, and ensure geometric consistency across views.
\begin{equation}
\label{Eq:loss}
\begin{aligned}
L_{RGB} &= (1-\beta)L_{MAE}(v_{gt}, v_{pred}) + \beta L_{SSIM}(v_{gt}, v_{pred}) \\
&\quad + \left( 1 - L_{NCC}\big(v_r(h_r), v_n(H_{rm} h_r)\big) \right),
\end{aligned}
\end{equation}
where $\beta$ is the balance weight. 
$v_{gt}$ and $v_{pred}$ denote the ground-truth and rendered views, respectively. 
$v_r$ and $v_n$ represent the reference and target views. 
$h_r$ is the homogeneous pixel coordinate in the reference view, 
and $H_{rm}$ is the corresponding homography matrix.
\\
\textbf{Signed Distance Inference for 3D Gaussians.} As described in Sec. 3.1, unlike previous methods that fit an implicit field from Gaussian points, we predict the signed distance for each Gaussian point $q_i = [x_i, y_i, z_i]$ using the implicit prior $f^t$. We efficiently interpolate the distance using trilinear interpolation,
\begin{equation}
\label{Eq:rgbloss}
\begin{aligned}
f^{t}(q_i) &= a_0 + a_1 x_i + a_2 y_i + a_3 z_i + a_4 x_iy_i \\&+ a_5 x_iz_i + a_6 y_iz_i + a_7 x_iy_iz_i ,
\end{aligned}
\end{equation}
where $\{a_0, ..., a_7\}$ are the interpolation weights of the TSDF grid vertices surrounding $q_i$.
% {插值3D高斯。}
\section{More Comparisons and Results}
In this subsection, we conduct additional ablation studies on the DTU dataset under the default parameter settings and provide more visual results and comparisons on both synthetic and large-scale scene datasets.

\begin{figure*}[!t]
    \centering
    \includegraphics[width=\textwidth]{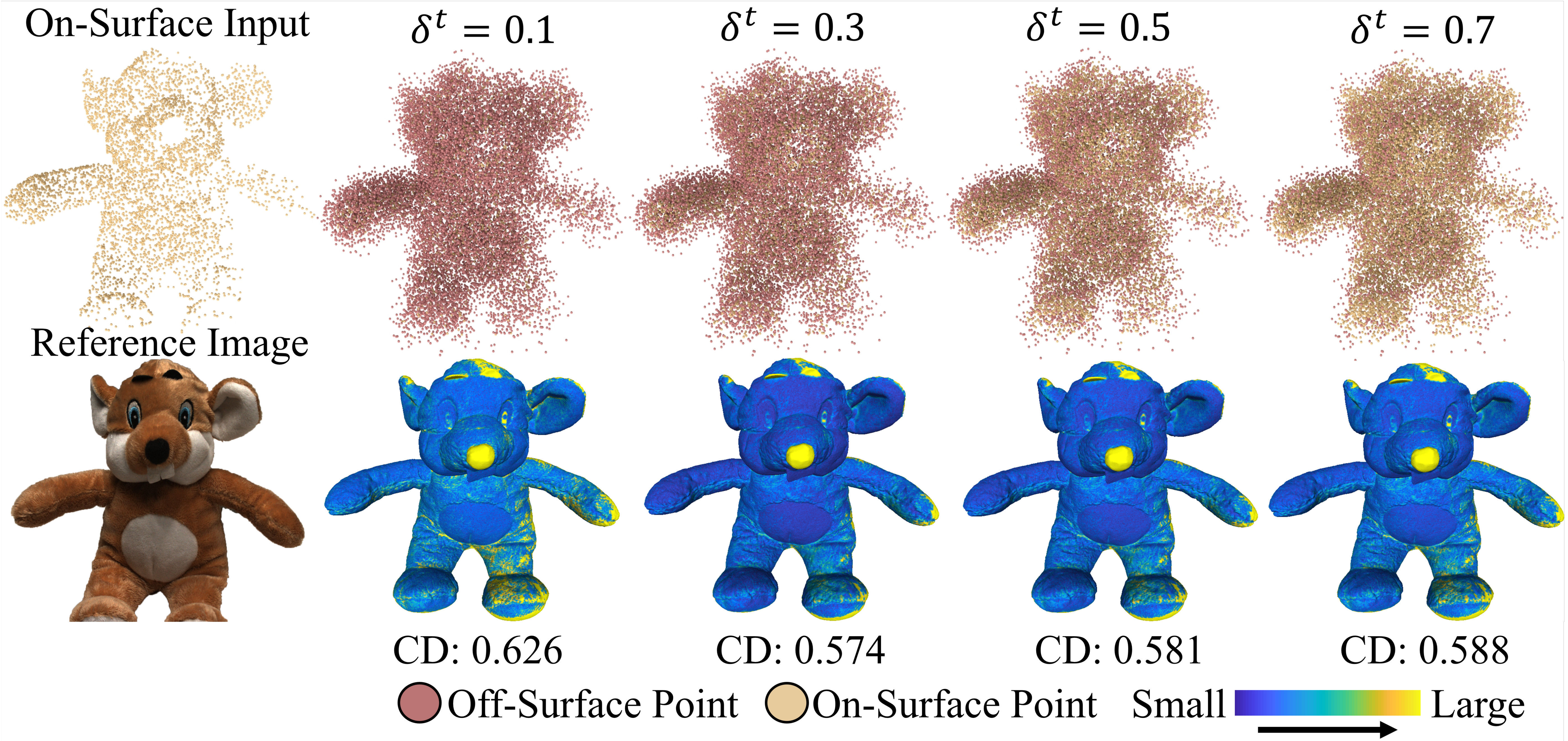}
    \caption{Effect of Opacity Constraint Threshold.}
    % \vspace{-0.2in}
    \label{fig:supp_ab_delta}
\end{figure*}

\noindent\textbf{Effect of Opacity Constraint Threshold. }We analyze the effect of the opacity constraint threshold $\delta^t$ on splitting 3D Gaussians into on-surface and off-surface subsets. Specifically, we add noise to surface points to evaluate the accuracy of $f^{t}$ under different thresholds. We  set $\delta^t$ to 0.7, 0.5, 0.3, and 0.1, respectively. As shown in Fig.~\ref{fig:supp_ab_delta}, a lower value of $\delta^t$ makes Gaussians more sensitive to be classified as off-surface points, while the higher ones relax the threshold but decreases the supervision accuracy. We further visualize the Gaussian classification results of $f^t$ under different settings. When $\delta^t$ is set to 0.1, most on-surface points are incorrectly labeled as outliers. In contrast, $f^t$ tends to label outliers as surface points as $\delta^t$ is higher than 0.3. To balance robustness and accuracy, we set $\delta^t$ to 0.3 by default.

\begin{figure*}
    \centering
    \includegraphics[width=\textwidth]{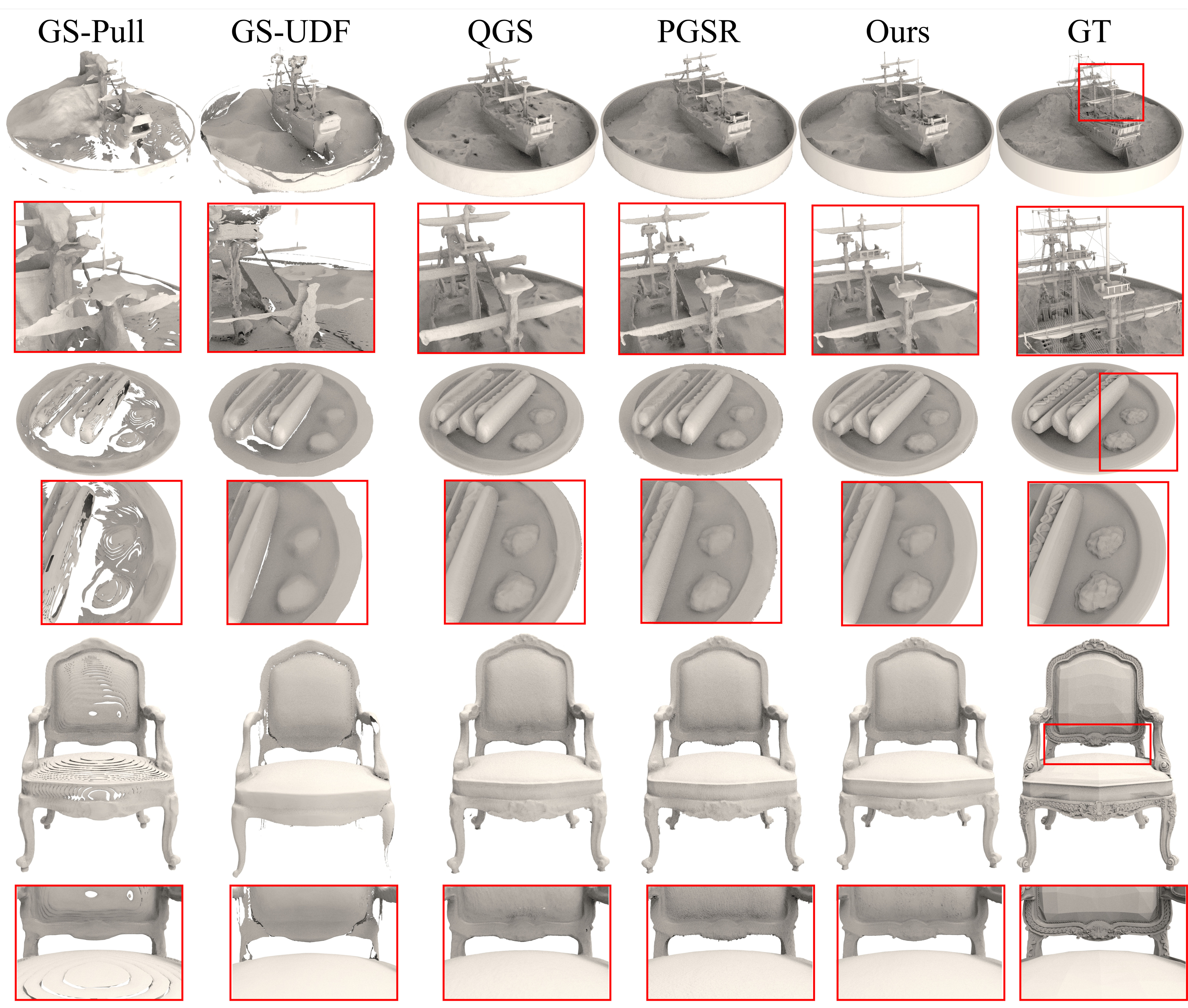}
    \caption{Visual Comparisons on NeRF-Synthetic Dataset.}
    % \vspace{-0.2in}
    \label{fig:supp_nerf-recon}
\end{figure*}

\noindent\textbf{Visual Comparisons on NeRF-Synthetic. } We provide additional comparisons in Fig. \ref{fig:supp_nerf-recon}. compared with implicit-field-based approaches (GS-Pull~\cite{zhang2024gspull} and GS-UDF~\cite{li2025gaussianudf}), our method reconstructs more complete surfaces in open-surface scenarios. In contrast to TSDF based methods (QGS~\cite{zhang2025quadratic} and PGSR~\cite{chen2024pgsr}), our approach leverages geometric priors to better preserve edge structures and reduce outliers and truncation artifacts on the boat hull or plate. Our method also recovers local details in scenes with complex textures. We further provide serial rendering visualization and evaluation in our video.

\begin{figure*}
    \centering
    \includegraphics[width=.9\textwidth]{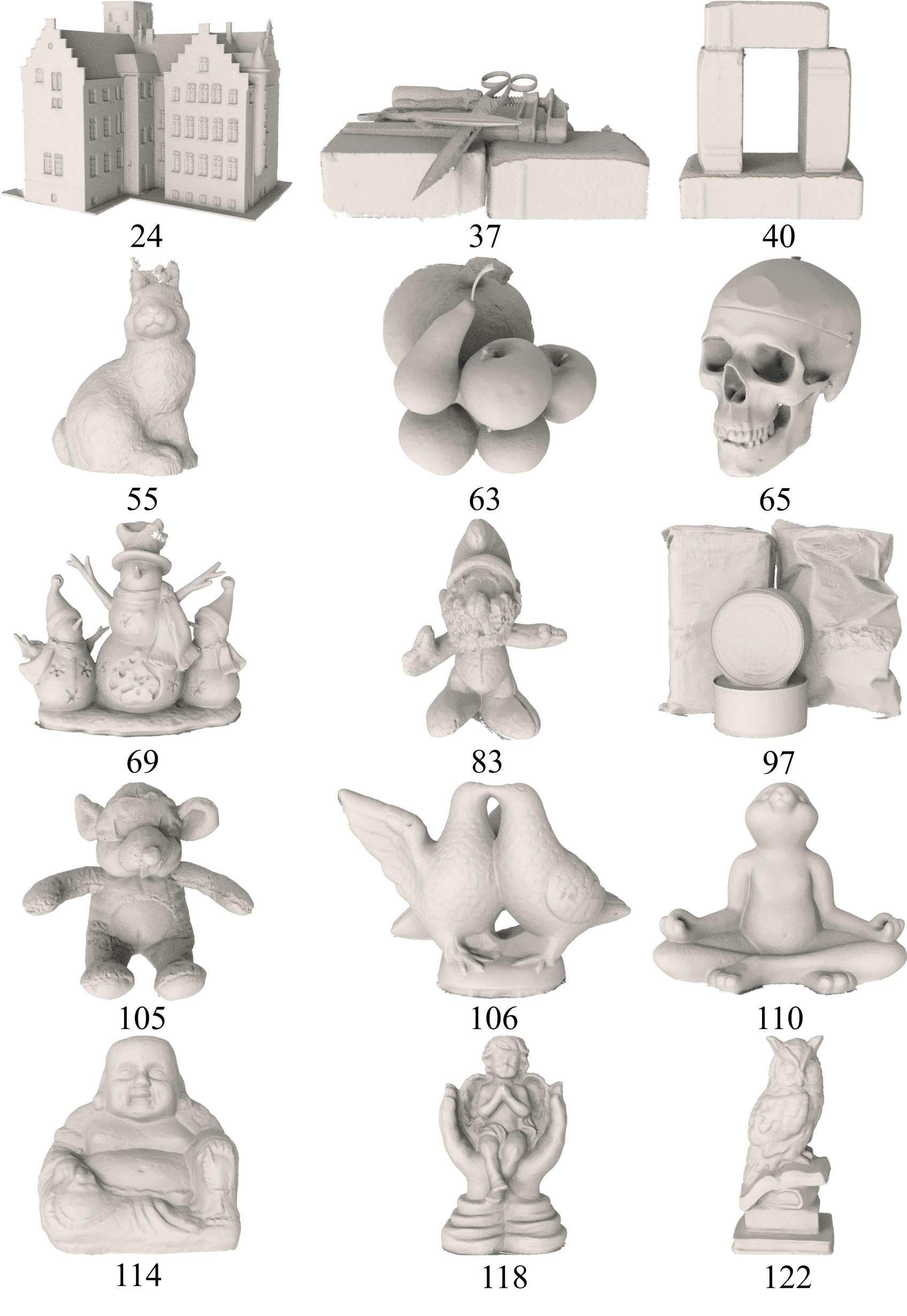}
    \caption{Visual Comparisons on DTU Dataset.}
    \label{fig:supp_dtu-recon}
\end{figure*}

\noindent\textbf{Visualization on DTU. }We also present qualitative results on the DTU dataset in Fig.~\ref{fig:supp_dtu-recon}. As shown in Fig.~\ref{fig:supp_dtu-recon}, our method recovers more accurate surfaces in regions with complex geometric topology and maintains consistency across various open scenes. 

\begin{figure*}
    \centering
    \includegraphics[width=\textwidth]{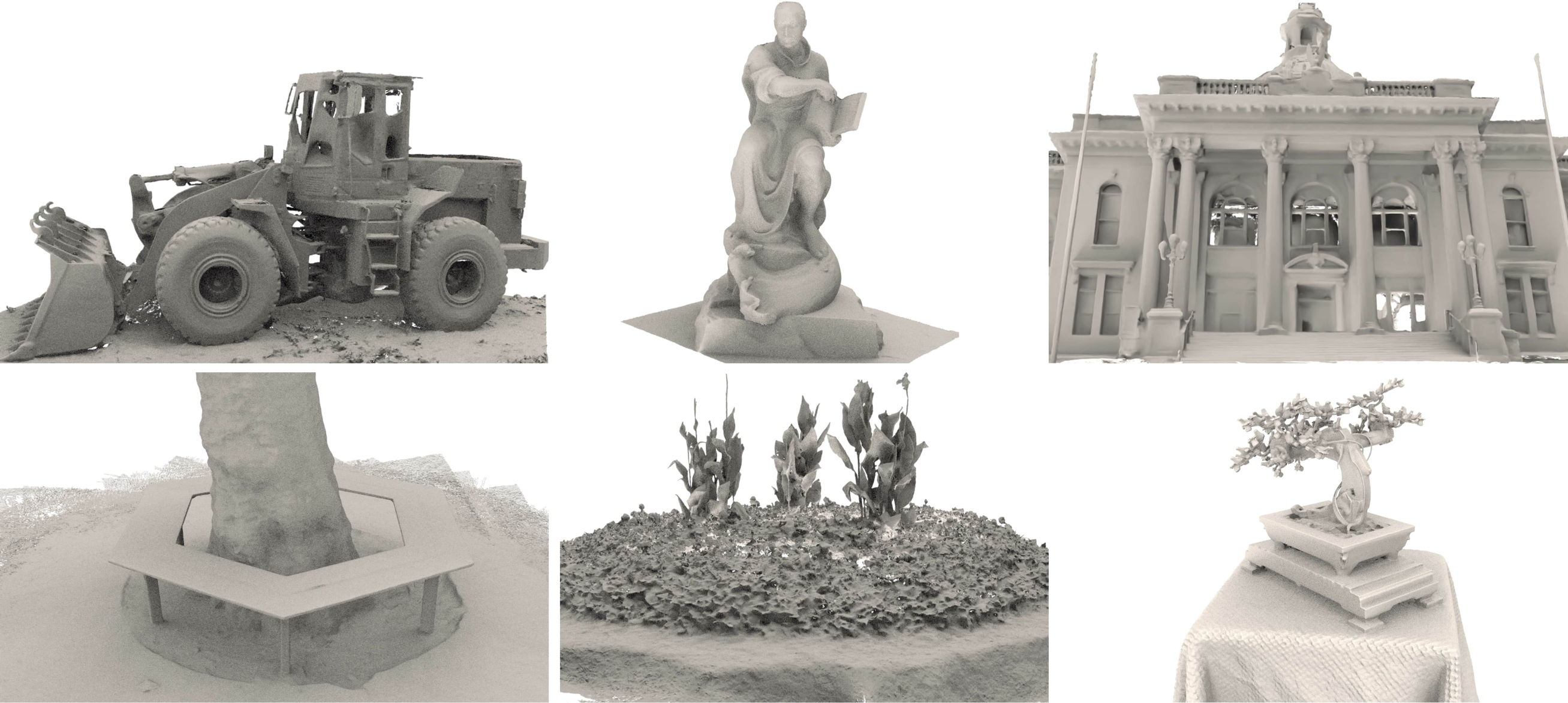}
    \caption{Surface Reconstruction Visualization on TNT and Mip-NeRF 360 Dataset.}
    \label{fig:supp_dtu-mip-recon}
\end{figure*}

\begin{figure*}
    \centering
    \includegraphics[width=\textwidth]{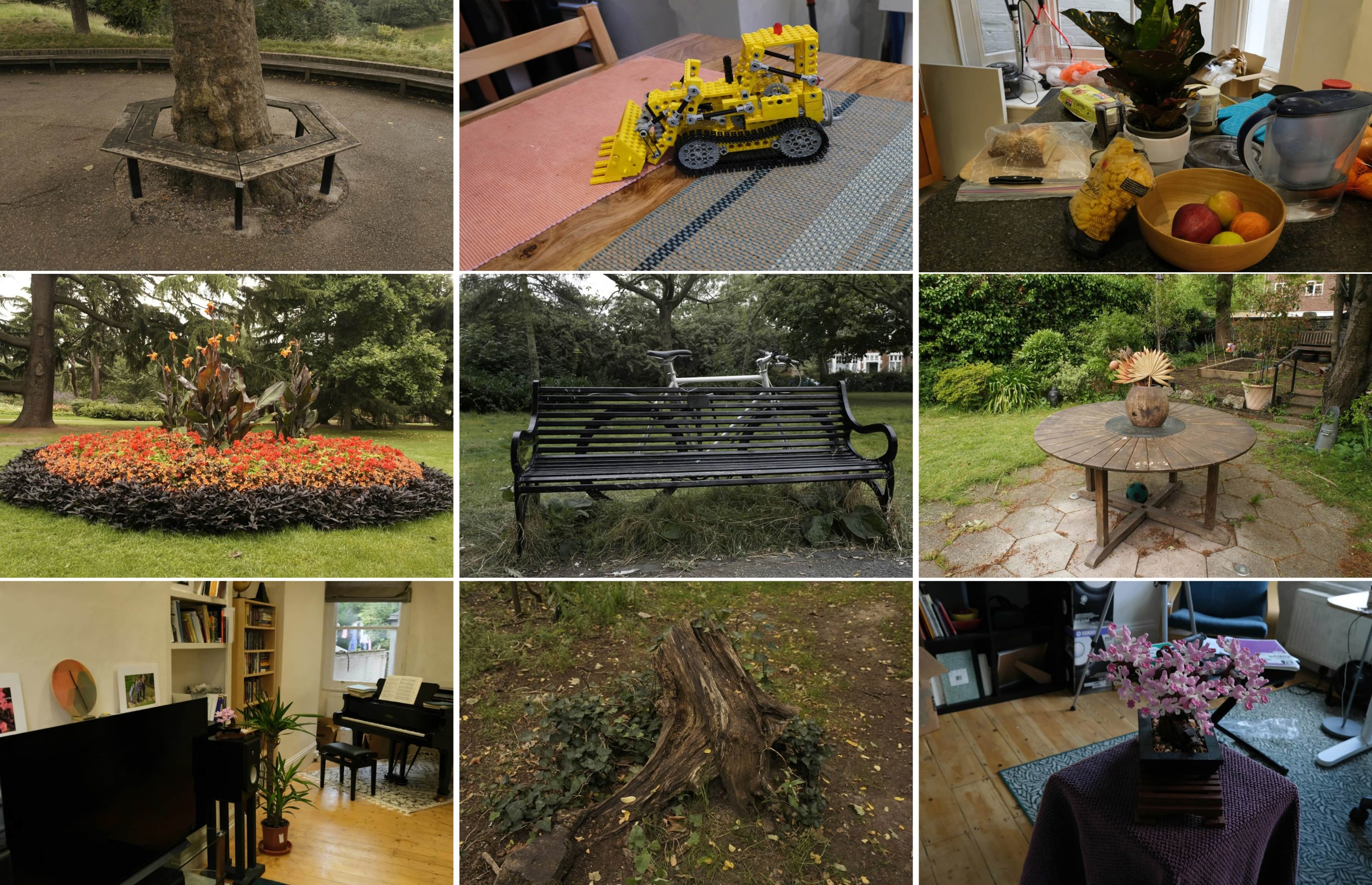}
    \caption{Rendering Visualization on Mip-NeRF 360 Dataset.}
    \label{fig:supp_render-mip}
\end{figure*}

\noindent\textbf{Visualization on TNT and Mip-NeRF 360. }We further validate the geometry inference and novel view rendering performance for our method on large-scale scenes (TNT and Mip-NeRF 360) in Fig.~\ref{fig:supp_dtu-mip-recon} and Fig.~\ref{fig:supp_render-mip}. As shown in Fig. ~\ref{fig:supp_dtu-mip-recon}, our method recovers accurate geometric structures in both complex indoor and outdoor environments. Moreover, we produce high-fidelity rendered views under low-texture regions and challenging lighting conditions in Fig.~\ref{fig:supp_render-mip}.
\section{Optimization Terms Explanation}
\textbf{Mean Absolute Error (MAE).}
SSIM evaluates the perceptual similarity between two views considering contrast and structure. Given the ground-truth view $v_{gt}$ and the predicted view $v_{pred}$,
the MAE loss computes the average absolute pixel-wise difference and is formulated as,
\begin{equation}
L_{\text{MAE}}(v_{gt}, v_{pred}) =
\lvert v_{gt} - v_{pred} \rvert .
\end{equation}
\textbf{Structural Similarity (SSIM).}
\noindent SSIM evaluates the perceptual similarity between two views considering
luminance, contrast, and structure. Given the ground-truth view $v_{gt}$ and the predicted view $v_{pred}$, the SSIM is defined as,
\begin{equation}
L_{SSIM}(v_{gt}, v_{pred}) =
\frac{(2\mu_{gt}\mu_{pred} + T_1)(2\sigma_{gt,pred} + T_2)}
     {(\mu_{gt}^2 + \mu_{pred}^2 + T_1)(\sigma_{gt}^2 + \sigma_{pred}^2 + T_2)},
\end{equation}
where $\mu_{gt}$ and $\mu_{pred}$ are means, $\sigma_{gt}^2$ and $\sigma_{pred}^2$ are the variances, $\sigma_{gt,pred}$ is the covariance between $v_{gt}$ and $v_{pred}$, $T_1$ and $T_2$ are constants.

\noindent\textbf{Normalized Cross-Correlation (NCC).}
NCC measures the linear correlation between two views after removing their mean intensities.
Given the ground-truth view $v_{gt}$ and the predicted view $v_{pred}$, the NCC is defined as,
\begin{equation}
L_{NCC}(v_{gt}, v_{pred}) =
\frac{\sum_{i=1}^{N} (v_{gt}^{(i)} - \mu_{gt})(v_{pred}^{(i)} - \mu_{pred})}
     {\sqrt{\sum_{i=1}^{N}(v_{gt}^{(i)} - \mu_{gt})^2}
      \sqrt{\sum_{i=1}^{N}(v_{pred}^{(i)} - \mu_{pred})^2}} ,
\end{equation}
where $i$ is the pixel index and $N$ is the total number of pixels. 
$\mu_{gt}$ and $\mu_{pred}$ denote the mean intensities of all pixels in $v_{gt}$ and $v_{pred}$, respectively.

% \section{Limitation and Future Works}
\section{Codes}
We provide a demonstration code as a part of our supplementary materials. We will release the source code and data upon acceptance.
\section{Video}
 We provide a video containing the visualizations on all datasets as a part of our supplementary materials.

\clearpage
% \label{sec:rationale}
% % 
% Having the supplementary compiled together with the main paper means that:
% % 
% \begin{itemize}
% \item The supplementary can back-reference sections of the main paper, for example, we can refer to \cref{sec:intro};
% \item The main paper can forward reference sub-sections within the supplementary explicitly (e.g. referring to a particular experiment); 
% \item When submitted to arXiv, the supplementary will already included at the end of the paper.
% \end{itemize}
% % 
% To split the supplementary pages from the main paper, you can use \href{https://support.apple.com/en-ca/guide/preview/prvw11793/mac#:~:text=Delete%20a%20page%20from%20a,or%20choose%20Edit%20%3E%20Delete).}{Preview (on macOS)}, \href{https://www.adobe.com/acrobat/how-to/delete-pages-from-pdf.html#:~:text=Choose%20%E2%80%9CTools%E2%80%9D%20%3E%20%E2%80%9COrganize,or%20pages%20from%20the%20file.}{Adobe Acrobat} (on all OSs), as well as \href{https://superuser.com/questions/517986/is-it-possible-to-delete-some-pages-of-a-pdf-document}{command line tools}.

\newpage
\quad
\newpage
\quad
\newpage
\quad
\newpage
\quad

%\clearpage

% WARNING: do not forget to delete the supplementary pages from your submission 
%\input{sec/X_suppl}

\end{document}